\documentclass{article}
\usepackage{amssymb}

\usepackage[final]{corl_2025} % Uncomment for the camera-ready ``final'' version.

\usepackage{amsmath,amssymb,amsfonts, amsthm}
\usepackage{graphicx}
\usepackage{wrapfig}
\usepackage[font=scriptsize,labelsep=period]{caption}
\usepackage[font=scriptsize,labelsep=period]{subcaption}
\usepackage{subfiles}
\usepackage[capitalize]{cleveref}
\usepackage{multicol}
\usepackage{multirow}
\usepackage{bbold}
\usepackage{custom}

\usepackage{todonotes}

\title{Learning Smooth State-Dependent Traversability from Dense Point Clouds}
\author{
\textbf{Zihao Dong\textsuperscript{1}},
\textbf{Alan Papalia\textsuperscript{1, 2}},
\textbf{Leonard Jung\textsuperscript{1}},
\textbf{Alenna Spiro\textsuperscript{1}}, \\
\textbf{Philip R. Osteen\textsuperscript{3*}},
\textbf{Christa S. Robison\textsuperscript{3*}},
\textbf{Michael Everett\textsuperscript{1}} \\
\textsuperscript{1}Northeastern University \quad
\textsuperscript{2}University of Michigan, Ann Arbor \\
\textsuperscript{3}DEVCOM Army Research Laboratory (ARL) \quad
\textsuperscript{*}Equal Contribution
}

\begin{document}
\maketitle

%===============================================================================

\begin{abstract}
    A key open challenge in off-road autonomy is that the traversability of terrain often depends on the vehicle's state.
    In particular, some obstacles are only traversable from some orientations.
    However, learning this interaction by encoding the angle of approach as a model input demands a 
    large and diverse training dataset and is computationally inefficient during planning due to repeated model inference.
    To address these challenges, we present SPARTA, a method for estimating approach angle conditioned 
    traversability from point clouds.
    Specifically, we impose geometric structure into our network by outputting a smooth analytical function 
    over the 1-Sphere that predicts risk distribution for any angle of approach 
    with minimal overhead and can be reused for subsequent queries.
    The function is composed of Fourier basis functions, which has important advantages for generalization due to their periodic nature
    and smoothness.
    We demonstrate SPARTA both in a high-fidelity simulation platform, where our model achieves a 91\% success rate crossing
    a 40m boulder field (compared to 73\% for the baseline), and on hardware, illustrating the generalization ability of the model to 
    real-world settings.
    Our code will be available at \url{https://github.com/neu-autonomy/SPARTA}.
    
\end{abstract}
% \blfootnote{Distribution Statement A: Approved for public release; distribution is unlimited.}
%===============================================================================

% Submission to CoRL 2025 will be entirely electronic, via a web site (not email). Information about the submission process and \LaTeX{} templates are available on the conference web site at \url{https://corl.org/}. For camera ready submission, use the \texttt{final} option for the \verb|\usepackage| command.

\section{Introduction}

\def\stateFootnote{\footnote{
        We focus on angle-dependency in this paper,
        and leave the extension to other state variables to future work.
    }}
\begin{wrapfigure}[13]{r}{0.45\textwidth}
    \centering
    \vspace{-1.2em}
    \includegraphics[width=\linewidth]{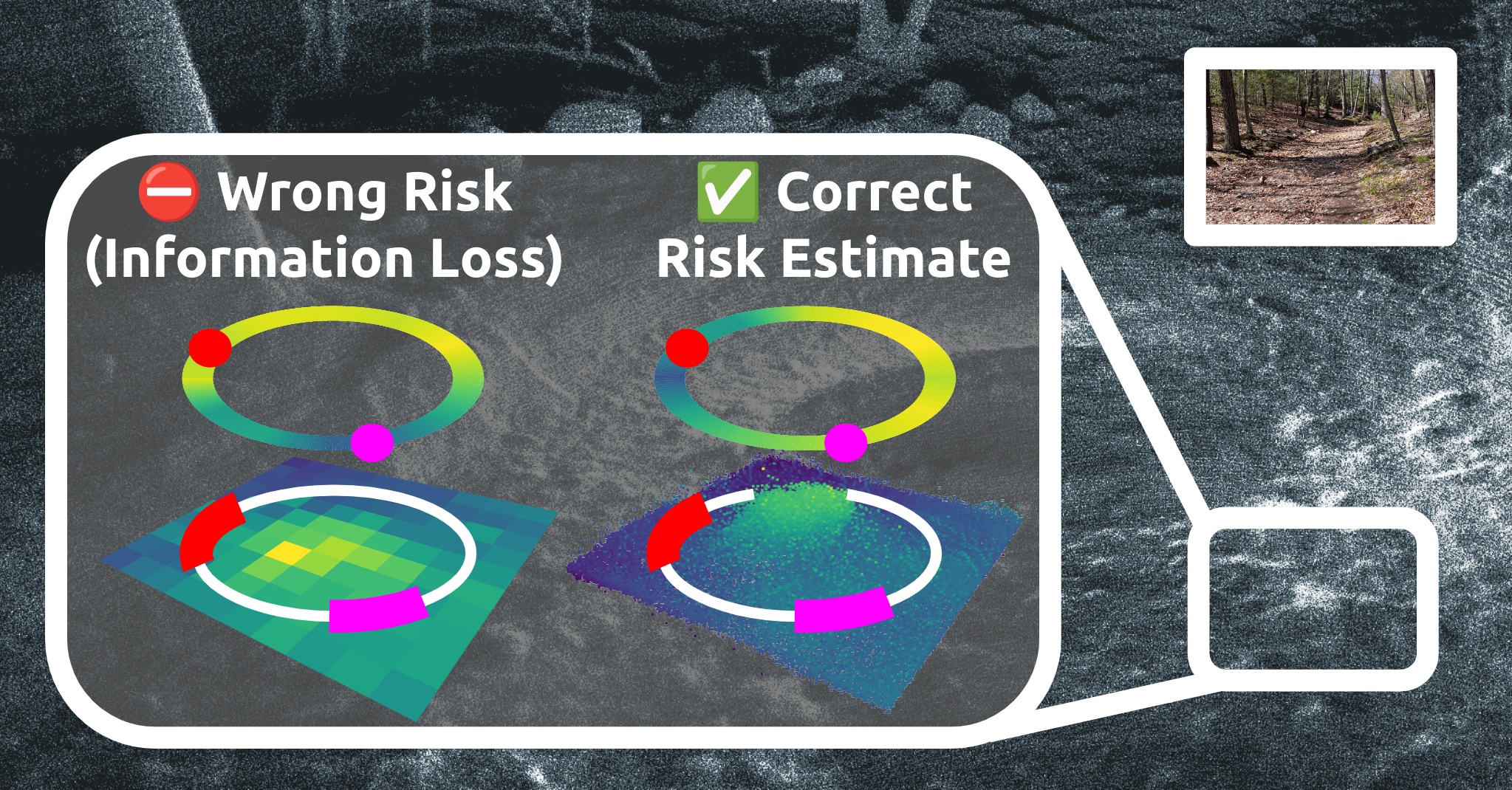}
    \caption{Dense point cloud captures geometric detail of the environment (Top right), while discretizing it into
        elevation map results in information loss. The underlying terrain is clearly distinguishable from noise (leaves, sensor noise) in
        the point cloud, but doing so is hard in an elevation map, leading to wrong risk estimation (Viridis Circle where Green denotes low risk).}
    \label{fig:pcd}
\end{wrapfigure}
% 1) what do we want to do and why does it matter?
Many applications for autonomous mobile robots involve traversing rough, off-road
environments~\citep{oliveira2021advances, lluvia2021active, wang2023development}.
%
% One key to reliably navigating in these environments is 
One key to reliably navigate in these environments is to estimate terrain traversability, i.e.,
the ability to understand the interaction between the robot and the terrain, and assess relevant
risk metrics (e.g., potential vehicle damage).
% 2) why is this hard?
Despite recent success in identifying traversable
terrain~\citep{cai2024evora, frey2023fast, meng2023terrainnet, frey2024roadrunner},
motion planning in geometrically complex environments (e.g., dense forests, boulder fields) remains a challenge.
Specifically, the risk associated with traversing a given terrain
is often a complicated function of the robot's state, for example its speed and
angle of approach, and the terrain geometry\stateFootnote.
% \blfootnote{\red{Our code will be available at \url{https://github.com/neu-autonomy/SPARTA}.}}.
%
% While high-speed traversal through risky terrain can be explicitly penalized
% within the planner, it is less 
It is especially unclear how to account for the angle of approach (e.g., using hand-crafted risk heuristics based on
terrain elevation~\citep{fan2021step, datar2024learning, lee2025trg}),
as this requires both detailed terrain information and nuanced consideration of
the underlying geometry.
%
% While detailed terrain representations are accessible and mature (e.g., point
% clouds), it is not apparent how to construct handcrafted risk heuristics that
% can properly account for the diverse, complex set of terrain we may expect
% to encounter in the wild.

% 3) our approach and how does it address these limitations?
\def\fourierFootnote{\footnote{The choice of a Fourier basis has
        deep theoretical roots connected to harmonic analysis and representation
        theory over the 1-sphere $S^1$~\citep{}.}}
In contrast, learning-based approaches have promise in being able to flexibly model
the complex relationship between terrain and traversal risk.
In particular, point cloud-based approaches are suitable for the problem setting,
because point clouds are capable of capturing fine geometric details of the environment
that would be missed in other representations like an elevation map.
As shown in \cref{fig:pcd}, in the point cloud, the terrain (ground) is distinguishable from
noise (e.g. leaves).
As a result, a point cloud-based method would prefer the safe approach angle (Red) over the
risky one (Magenta) due to a relatively smooth elevation change.
However, noise and terrain cannot be easily separated in the elevation map despite its high
resolution (5cm).
The terrain appears smoother from the risky angle (Magenta) thus an elevation map-based
method would (incorrectly) select it instead.

In this paper, we propose a technique to learn risk variables of interest,
based on a point cloud representation of the terrain.
Specifically, to model system noise implicitly captured in the dataset,
we represent risk as a categorical distribution due to its ability
to approximate arbitrary distributions~\citep{cai2024evora}.
% \alanInline{ZIHAO PLZ FINISH THIS PLZ: we want to provide some explanation of
%     why our approach (a bounded distribution approximated by a categorical
%     distribution) is a reasonable thing to do. I think it's okay if we have a
%     long chain of thought right now and explictly spell things out. Once we have
%     a clear explanation of how and why we are choosing to describe risk this
%     way, we can go back and make it more concise.}
% \zihaoInline{I rephrased the motivation to emphasize the connection to 
% better generalization and data-efficiency, over efficiency querying during planning.}
%
However, a point cloud-based network that naively encodes the angle of approach
would be hard to train in practice due to the sparsity of data in robotics applications.
As a result, our model takes in the point cloud and predicts an analytical function
(represented using Fourier basis functions) that maps angle of approach to the target risk variable distribution.
The choice of Fourier basis ensures that the resulting function is periodic, thus avoiding
the wrap-around discontinuity~\citep{zhou2019continuity}, and has important advantages
 for generalization in the training process due to its upper-bounded Lipschitz constant by construction.
The function, once computed for a terrain patch, can be queried efficiently with
minimal computation overhead for any angle of approach at any timestep when integrated with modern planners
like MPPI~\citep{williams2016aggressive, williams2017information}.
To summarize, in this paper we propose SPARTA (\textbf{S}mooth \textbf{P}oint-cloud
\textbf{A}pproach-angle \textbf{R}easoning for \textbf{T}errain \textbf{A}ssessment)
with the following key contributions:

\begin{itemize}
    \item A novel learning-based approach leveraging Fourier basis functions,
          which estimates angle-of-approach dependent traversal risk for complex
          terrains represented as point clouds.
    \item Theoretical analysis of the advantage of imposing smoothness by using Fourier basis functions,
          with connection to Lipschitz continuity and model generalization capability.
    \item Evaluations of the proposed pipeline through learning to predict potential
          vehicle damage, a risk variable prior works handle using handcrafted rules,
          with extensive results both in a high-fidelity simulator \citep{beamng_tech} and 
          on hardware, with integration into a MPPI planner~\cite{williams2016aggressive}.
          In \cref{sec:appendix:example} we discuss \textbf{SPARTA}'s generalization
          to other problem settings.
\end{itemize}

\section{Related Work}
% Our approach is a framework which uses point-cloud information to estimate risk
% in traversing a given terrain. Accordingly, we will discuss three lines of
% research closely related to our problem: traversability estimation, learning
% dynamics models, and machine learning on point-cloud data. As our work focuses
% on estimating risk, and not necessarily on the downstream task of planning
% risk-aware trajectories we do not discuss works in X, Y, Z.

% \alanInline{If we don't talk about learned dynamics models, let's just say it is
% very important and aligned in final objective (improved planning) but it is sufficiently different in the underlying
% assumptions. E.g., our approach is more in the philosophy that
%  we often have sufficiently good base dynamics models; in general we don't need more complicated dynamics models, we
%  just need to avoid certain regions which would cause failure. In theory, you could likely capture this with
% }

\paragraph{Traversability Estimation}
Traversability estimation aims to identify terrain suitable for robot navigation.
Traversability estimates typically rely on either manually designed features~\citep{fan2021step, lee2025trg} or 
are derived from learned vehicle movement patterns~\citep{cai2022risk, cai2024evora, meng2023terrainnet, frey2024roadrunner, frey2023fast}.
Both~\citep{fan2021step} and~\citep{lee2025trg} use local elevation measurements to estimate
stepping difficulty using heuristics such as step-wise elevation change.
% %
% However, the traversal risk associated with a given terrain is a complicated function of both the
% robot's state and the detailed geometry, which is hard to account for using handcrafted heuristics.
% %
On the other hand, learned methods usually rely on multiple sensing modalities, especially
elevation maps and semantic maps~\citep{cai2022risk, cai2024evora, meng2023terrainnet, frey2024roadrunner, frey2023fast}.
WVN~\citep{frey2023fast} uses camera input and trains a traversable terrain classifier online leveraging pretrained features
from DINO-ViT~\citep{caron2021emerging}.
Meanwhile, \citep{meng2023terrainnet, frey2024roadrunner} fuse semantic and elevation measurements, and use separate
decoder heads to predict dense semantic and elevation maps, which are used to score planner rollouts using various heuristics.
\citep{cai2022risk, cai2024evora} use semantic and elevation maps to predict traversability distributions represented
as linear/angular traction and incorporate the worst-case traction into the vehicle dynamics model for risk-aware planning.
% %
% However, none of these approaches models the angle-dependent nature of traversability.
% %
% As a result, they can be overly conservative in geometrically complex environments, while underestimate traversal 
% risk when the terrain is only dangerous from a few orientations.
Several works modeled the state-dependency of traversability by encoding robot
 state as part of the neural networks' input~\citep{frey2022locomotion, weerakoon2022terp, seo2023scate}.
 However, there performances suffer from data sparsity and the dimensionality mismatch between exteroceptive inputs
 and low dimensional state space.

\paragraph{Learning Harmonic Functions}
Geometric deep learning has found its place in numerous machine learning applications.
For example, FBM~\citep{yang2024rethinking} uses Fourier basis functions to composite long-time series data
and achieved state-of-the-art performance on multiple benchmarks.
NeuRBF~\citep{chen2023neurbf} replaces traditional grid-based structures with radial basis functions for constructing implicit 
neural representation of scenes, allowing the model to better adapt to the underlying signal
structures.
In robotics, OrbitGrasp~\citep{hu2024orbitgrasp} leverages spherical harmonics functions to represent
a grasping quality function.
%
% We observe the geometric structure embedded in the traversability estimation problem, and as a result
% propose to leverage Fourier basis functions to exploit this intrinsic prior.
Off-road autonomy, especially traversability estimation, is a setting with rich geometric structure, 
and as a result we propose to leverage Fourier basis functions to exploit this intrinsic prior.

\paragraph{Placement of This Work}
% This work is most closely related to traversability estimation, yet it is complementary to existing methods.
% %
% We do not impose assumptions on the risk variable of interest, except the mild assumptions that it
% has clear lower and upper bounds and does not change abruptly with a small change in angle of approach.
% %
% As a result, existing works for estimating traversability (a risk variable) can be expanded with our method
% to achieve more granular predictions based on the robot's angle of approach.
% %
% This work is most closely related to traversability estimation.
Unlike previous traversability estimation algorithms, SPARTA explicitly models the angle dependency of risk variables 
by exploiting the geometric structure of angle of approach (the 1-Sphere).
Because we impose few assumptions on the risk variable of interest (i.e., it
has clear lower and upper bounds and does not change abruptly with a small change in angle of approach), 
our approach could be used to extend (most) existing traversability estimation methods for more accurate 
and granular prediction.
% %
% Most importantly, our formulation exploiting the geometric structure is general and finds application in other problem settings.
%
Besides tire deformation (demonstrated in our experiments \cref{sec:exp}), 
we show examples of variables with a similar geometric structure in \cref{sec:appendix:example}.

\section{Method}
% \alanInline{Let's start this section by describing the different components of what will be talked about and
% how they connect together. It seems like we are going to talk about:
% \begin{itemize}
%     \item how we represent risk
%     \item how we capture aleatoric uncertainty in our problem
%     \item how we \emph{by construction} have a continuous representation
%     \begin{itemize}
%         \item the way we construct this representation and it's connection to representation theory and irreducible representations
%         \item the theoretical advantages it has for learning efficiency
%     \end{itemize}
%     \item how we integrate our representation into a state-of-the-art planner (MPPI)
% \end{itemize}
% }

% \begin{itemize}
%     \item Explain using fourier basis functions, and a network to learn fourier coefficients, to construct signal on 1-Sphere (somehow I think we should not mention using tire deformation in this section, but rather save it to implementation detail or experiment setup)
%     \item Network architecture (brief)
%     \item Inference pipeline, including querying 1-sphere signals and getting the CVaR of the distribution
% \end{itemize}

\begin{figure}[t]
    \centering
    \vspace{-1.5em}
    \includegraphics[width=0.85\linewidth]{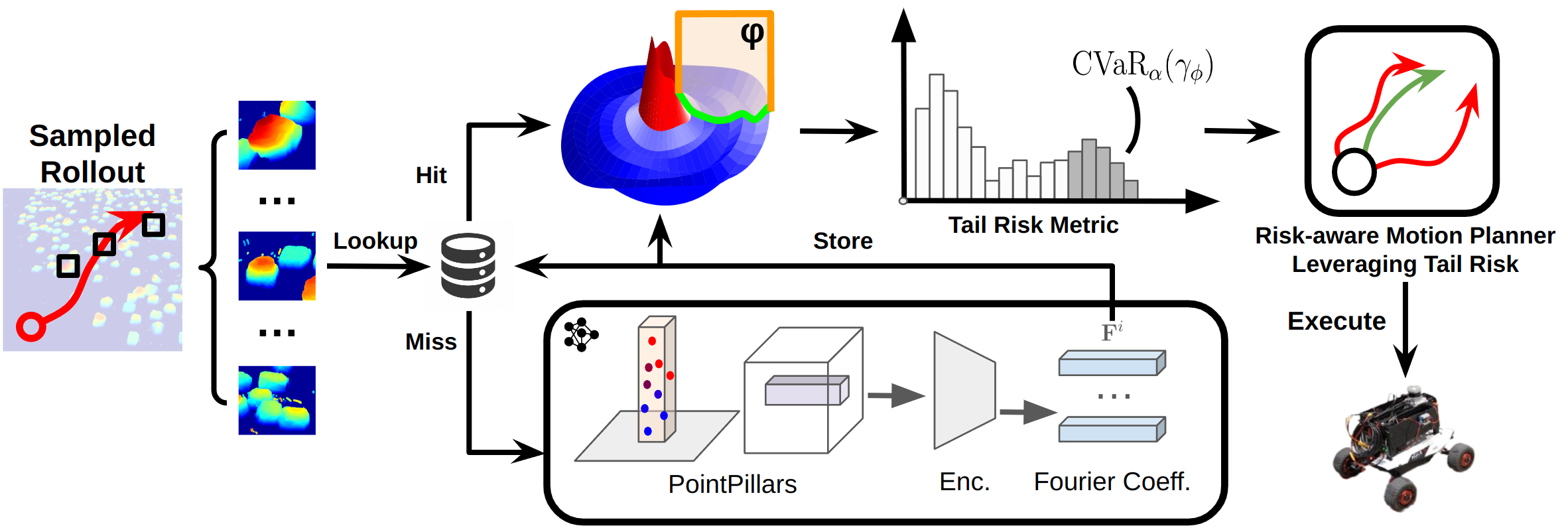}
    \caption{
    Overall pipeline of the proposed algorithm.
    For all terrain patches we are interested in, if its Fourier coefficients are not in the database, 
    we compute them and store in the database.
    Otherwise we retrieve the Fourier coefficients from the database and construct the analytical functions,
    which are queried to compute the risk variable distribution, whose tail distribution is considered for
    risk aware planning.
    % \alanInline{The trickiest part of our approach is that for each bin there is
    %     a separate periodic function that we are `learning'. We should make sure
    %     that our figure communicates this.}
    }
    \label{fig:method:pipeline}
    \vspace{-1.7em}
\end{figure}

In this section we introduce SPARTA, a method for predicting the risk
associated with traversing a terrain patch at a given angle based on prior data
traversing a wide variety of terrains (represented as dense point cloud).
One challenge in traversability estimation is handling aleatoric uncertainty,
the inherent uncertainty in a system which cannot be reduced with more
data~\citep{hullermeier2021aleatoric}.
SPARTA captures
this uncertainty by treating risk as a random variable -- a categorical
distribution with $B$ discrete bins -- whose concentration parameters are
predicted by our model (\cref{sec:method:uncertainty} and \cref{sec:method:representation}). 
The tail of the distribution is used to estimate risk for risk-aware planning (\cref{sec:method:planner}).
% \alan{Last sentence could be improved -- should say CVaR is std measure of risk}

The primary novelty in SPARTA is in representing this risk distribution.
SPARTA predicts an analytical function for each bin, and each function smoothly maps
any angle of approach to the corresponding risk concentration parameter
(\cref{sec:method:representation}).
We characterize the smoothness of these risk functions (\cref{sec:method:theory}),
which we hypothesize to be useful for generalization across approach angles (empirically supported in
\cref{sec:exp:performance}).
Once computed, the function can be re-used for subsequent queries for the same
terrain patch, thus avoiding repeated neural network inference during planning.
We include an experiment demonstrating the query efficiency of our model in 
\cref{sec:appendix:runtime}.
An overview of the risk estimation pipeline and its application to planning is
shown in \cref{fig:method:pipeline}.

\subsection{Modeling Aleatoric Uncertainty with Categorical Distributions}\label{sec:method:uncertainty}
% \alan{This section was a bit difficult to read, see attached comment.}
%
% \alanInline{
%     Start the section by saying what we want to show in this section. I.e.,
%     why is aleatoric uncertainty important and how do we capture it? It's good if
%     you can provide some high-level details for your approach here so that when
%     we start getting the more specific descriptions of the approach, the reader
%     can connect the low-level and high-level descriptions to get a better
%     intuition.
% }

In this work, we are interested in predicting a risk variable conditioned on 
the angle of approach.
However, the aleatoric uncertainty inherent to the training data will likely prevent us from 
reliably estimating risk as a single scalar value.
%
% We assume we have a dataset recording the risk variable while traversing various terrains.
% %
% However, the aleatoric uncertainty of this data will likely prevent us from reliably
% estimating risk as a single scalar value.
%
To develop robust real-world systems, it is important to properly account for aleatoric
uncertainty~\citep{cai2022risk, cai2024evora}.
SPARTA captures aleatoric uncertainty by modeling risk as a random variable
whose distribution reflects this uncertainty. 
Without assuming a specific parametric form, we approximate this distribution with
a categorical distribution~\citep{cai2024evora}.

More formally, given a terrain patch represented as a point cloud $\mathbf{Q}
\in \mathbb{R}^{m \times 3}$ where $m > 0$ is the number of points in the point
cloud and the angle of approach $\phi \in [0, 2\pi]$, we are interested in
obtaining a function $\Gamma \colon (\mathbf{Q}, \phi) \mapsto
\mathbf{p}(\gamma_{\mathbf{Q}, \phi})$, where $\gamma_{\mathbf{Q}, \phi} \in
\mathbb{R}$ denotes a risk related variable\footnote{For example, risk related variables may
include vehicle damage, terrain traction, or rollover risk.} and
$\mathbf{p}(\gamma_{\mathbf{Q}, \phi})$ is the probability mass function (PMF)
of a categorical distribution with $B$ bins.\footnote{For conciseness, we omit
the subscript $\mathbf{Q}$ in the rest of this paper.}
The categorical distribution $\mathbf{p}(\gamma_\phi)$ is parameterized by non-negative
concentration parameters $\boldsymbol{\bar{\gamma}_\phi} = [\gamma^1_\phi,\, \ldots,\,
\gamma^B_\phi] \in \mathbb{R}^B_{\geq 0}$ and computed by normalizing $\bar{\gamma}_\phi$ to sum to 1:
\begin{equation} \label{eqn:PMF}
    \mathbf{p}(\gamma_\phi) = \frac{\boldsymbol{\bar{\gamma}_\phi}}{\sum_{b=1}^B \gamma^b_\phi}.
\end{equation}

To obtain this risk distribution $\mathbf{p}(\gamma_\phi)$, we can train a
neural network 
% $\Gamma_\theta \colon (\mathbf{Q}, \phi) \mapsto \boldsymbol{\bar{\gamma}_\phi}$ 
% with weights $\theta$ can be trained 
on a dataset of traversals on various terrains, $\{(\mathbf{Q}_k, \mathbf{y}_k,
\phi_k)\}_{k=1}^K$, where $\mathbf{y}_k$ denotes the ground truth (empirical) risk PMF.
The network is trained to minimize the squared Earth Mover's Distance ($\text{EMD}^2$):
\begin{equation}\label{eqn:EMD2}
    L^{\text{EMD}^2}(\mathbf{p}(\phi_k), \mathbf{y}_k) = ||\text{cs}(\mathbf{p}(\phi_k)) - \text{cs}(\mathbf{y}_k)||_2^2,
\end{equation}
where $\text{cs}(\cdot)$ is the cumulative sum operator.
However, training such a network for risk-aware planning poses two major
difficulties.
% Issues:
% (i) capturing the smooth and periodic nature of the set of angles [0, 2\pi]
% (ii) computational efficiency of querying the model to perform planning
The first issue is due to the structure of angles of approach $\phi$, i.e.,
there is a periodicity of $2\pi$ and we expect that small changes in $\phi$
should lead to small changes in the risk distribution. While a neural network
can approximate this structure, it typically requires a large and diverse dataset to
do so.
The second issue is the model's query efficiency, which
becomes a major concern for real-time planning.
In subsequent sections, we discuss how SPARTA addresses these issues by
leveraging Fourier basis functions to construct a smooth and periodic analytical function.

\subsection{Leveraging $S^1$ as a Continuous Representation of Approach Angle}\label{sec:method:representation}

\def\irrepFootnote{\footnote{
        More precisely, the Fourier basis functions
        forms a complete orthonormal basis for $L^2(S^1, \mathbb{R})$,
        and corresponds to the real-valued matrix coefficients of the one-dimensional irreducible
        unitary representations (characters) of the compact abelian group $S^1$.
    }
}

Here we introduce how SPARTA represents the risk concentration parameters
$\gamma_\phi$ as a function of the angle of approach $\phi$ via Fourier basis
functions.
This representation stems from two main insights: (i) the set of angles
$[0, 2\pi]$ can be mapped to the 1-sphere $S^1$ (i.e., the unit circle that lies
in the terrain ground plane and is centered at the intersection of the terrain
normal vector and the plane) and (ii) Fourier basis function is a fundamental 
way of representing arbitrary functions over $S^1$.

% % \paragraph{Mapping angles to the 1-sphere.}
% It is well-known that angles $\phi \in [0, 2\pi]$ can be mapped to points on the
% unit circle $S^1$. For our application, we consider the unit circle to lie in
% the terrain ground plane and to be centered at the intersection of the terrain
% normal vector and the plane.

% \paragraph{Fourier basis functions.}
\begin{wrapfigure}[10]{r}{0.4\textwidth}
    \centering
    \vspace{-1.0em}
    \begin{subfigure}[b]{0.16\textwidth}
        \centering
        \includegraphics[width=\linewidth]{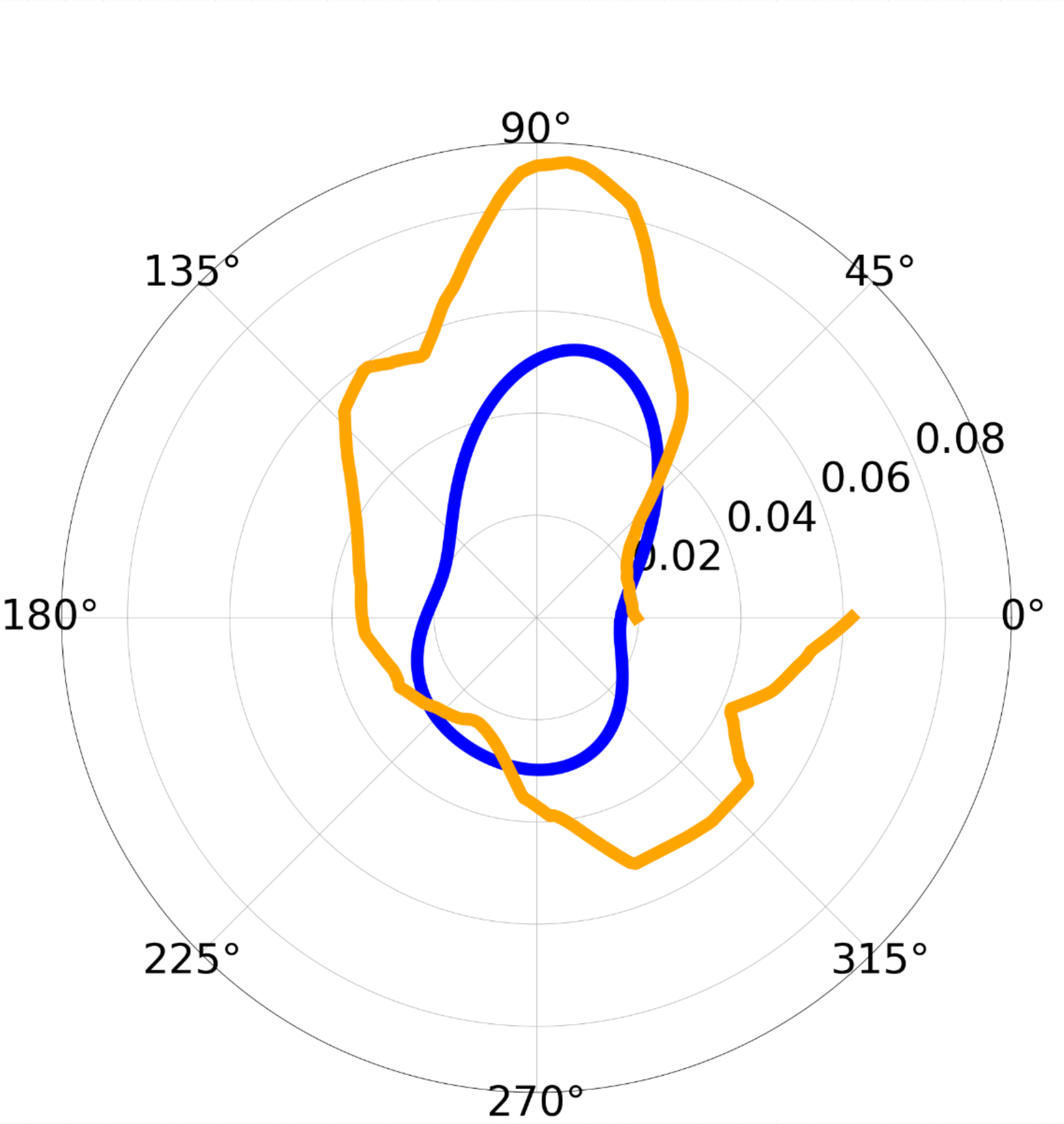}
        \caption{}
        \label{fig:method:wrap-around:a}
    \end{subfigure}
    \hspace{0.02\linewidth}
    \begin{subfigure}[b]{0.16\textwidth}
        \centering
        \includegraphics[width=\linewidth]{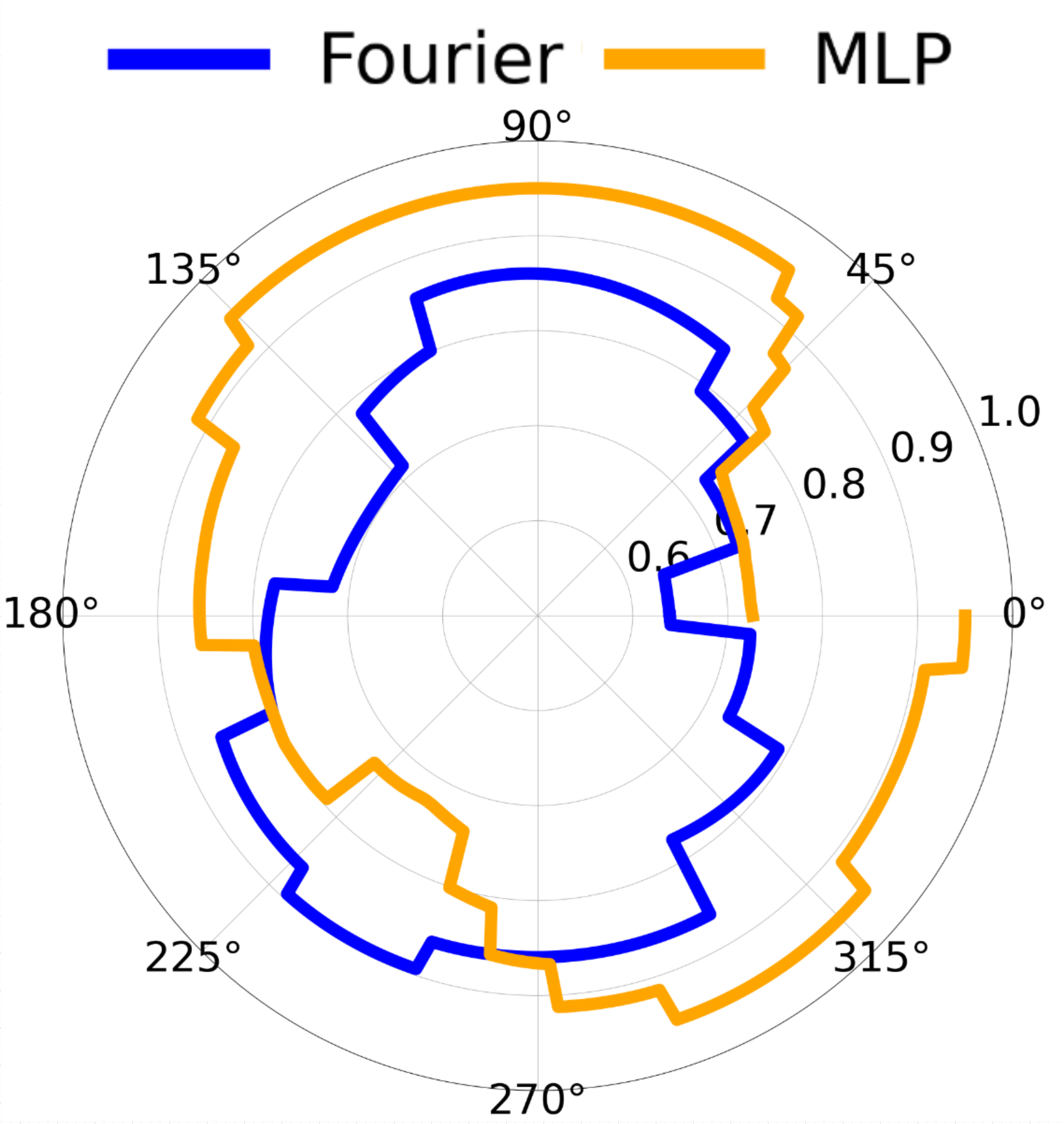}
        \caption{}
        \label{fig:method:wrap-around:b}
    \end{subfigure}
    % \captionsetup{font=footnotesize}
    \vspace{-5pt}
    \caption{
        (a): Naive model (orange) suffers from wrap-around
        discontinuity, whereas our model (blue) predicts continuous signal.
        (b): The phenomenon propagates to the CVaR of the predicted distribution.
    }
    \label{fig:method:wrap-around}
\end{wrapfigure}
Given a representation of angles on $S^1$, we can consider the risk
concentration parameters $\gamma_\phi$ as functions over $S^1$. It
follows from harmonic analysis and representation theory \cite{rudin2017fourier}
that any square-integrable real-valued function over the 1-sphere $S^1$ can be
uniquely decomposed into a (possibly infinite) linear combination of the Fourier
basis functions $\{1\} \cup \{\cos(k\phi), \sin(k\phi): k \in [1, n]\}$.
This suggests these basis functions as the natural choice for representing
arbitrary functions on the 1-sphere.
%\irrepFootnote
%
Altogether this allows us to compute risk concentration parameters at angle
$\phi$ by taking a linear combination of the Fourier basis functions
and applying the sigmoid function\footnote{
    While we used sigmoid function, other non-negative
    functions could theoretically be used (e.g., softplus).
    % We choose sigmoid because it is bounded and smooth.
}
$\sigma(\cdot)$ to ensure non-negativity (so
the result is a valid concentration parameter):
% through a sigmoid function $\sigma(\cdot)$ (to ensure non-negativity):
\begin{equation} \label{eqn:fourier}
    \gamma_\phi^i
    % f^i(\phi) 
    = \sigma(\frac{a_i^0}{2} + \sum_{k=1}^n [ a_i^k \cos(k\phi) + b_i^k \sin(k\phi) ]).
\end{equation}
%

% \begin{wrapfigure}[11]{r}{0.4\textwidth}
%     \centering
%     \vspace{-1.0em}
%     \begin{subfigure}[b]{0.19\textwidth}
%         \centering
%         \includegraphics[width=\linewidth]{images/method/wrap_around_bin_new_ratio.png}
%         \caption{}
%         \label{fig:method:wrap-around:a}
%     \end{subfigure}
%     \begin{subfigure}[b]{0.19\textwidth}
%         \centering
%         \includegraphics[width=\linewidth]{images/method/wrap_around_cvar_new_ratio.png}
%         \caption{}
%         \label{fig:method:wrap-around:b}
%     \end{subfigure}
%     % \captionsetup{font=footnotesize}
%     \vspace{-5pt}
%     \caption{
%         (a): Naive model (orange) suffers from wrap-around
%         discontinuity, whereas our model (blue) predicts continuous signal.
%         (b): The phenomenon propagates to the CVaR of the predicted distribution.
%     }
%     \label{fig:method:wrap-around}
% \end{wrapfigure}
%
Concretely, our neural network $\Gamma_\theta$ takes an input point cloud $\mathbf{Q}$ and, for
each bin $i \in [1, B]$, predicts the corresponding coefficients
$\mathbf{F}^i \triangleq [a_i^0, a_i^1, b_i^1, \cdots a_i^n, b_i^n ] \in
    \mathbb{R}^{2n+1}$.
\cref{eqn:fourier} is then used to compute the concentration parameters
$\boldsymbol{\bar{\gamma}_\phi}$ during planning.
\cref{eqn:fourier} is fast to compute as it requires a single (batched) dot
product and a pass through the sigmoid function, and its runtime scales linearly
with the maximum frequency $n$, a small number in practice\footnote{we found $n=3$ to
suffice in our experiments. We include model runtime comparison
in \cref{sec:appendix:runtime}.}.
We note that the choice of Fourier basis function naturally avoids
wrap-around discontinuities at $\phi=2\pi$. We demonstrate
this in \cref{fig:method:wrap-around:a}, where we compare the predictions
of a baseline multilayer perceptron model (orange) that encodes the angle of
approach $\phi$ directly as a model input, and our model (blue).
It can be seen that the baseline model (orange) is discontinuous at the wrap-around
point, resulting in a large jump in the predicted concentration parameters.
% predicting different concentration parameters.
%
This discontinuity is known to be harmful to model
training when working with angle-related
variables~\citep{zhou2019continuity}.
In contrast, our model (blue) naturally respects the periodicity and generates a
globally smooth signal over $S^1$.
Besides avoiding the wrap-around discontinuity, the Fourier basis functions allow
better generalization when trained with limited data, as we detail
through analyzing its Lipschitz constant in the next subsection.\par

\subsection{Imposed Smoothness Improves Prediction Generalization}\label{sec:method:theory}

\def\dataFootnote{\footnote{Previous work observed that for
the bound on a neural network's Lipschitz constant to decrease, both
sufficient model capacity and large amounts of training data ($\approx$ number of model parameters)
are crucial~\citep{belkin2019reconciling}.}}

In off-road settings we may expect that small changes in approach angle $\phi$
induce relatively small changes in the risk variable $\gamma_\phi$.
In this subsection we show that $\gamma_\phi$ is smooth with respect to $\phi$
and provide a simple upper bound on the first derivative (i.e., an upper bound
on the Lipschitz constant).
% which is tight under easily identifiable conditions.
%
We connect this Lipschitz constant analysis to results in learning theory
\citep{fazlyab2019efficient, khromov2023some} to motivate the hypothesis that
our representation improves generalization  (see \cref{sec:exp:performance} for
empirical evidence).

Formally, if $L^i$ denotes the Lipschitz constant of the mapping from approach
angle to risk concentration parameter
$\phi \rightarrow \gamma_\phi^i$, then adding a small perturbation $\delta$
to $\phi$ bounds the change in $\gamma_\phi^i$: $| \gamma_{\phi+\delta}^i - \gamma_\phi^i | \leq L^i | \delta |$.
% \begin{equation} \label{eqn:lipschitz}
%     | \gamma_{\phi+\delta}^i - \gamma_\phi^i | \leq L^i | \delta |.
% \end{equation}
%
% %
% Ideally, the Lipschitz constant $L^i$ is small.
% %
% In a naive approach of explicitly encoding the angle of approach $\phi$, we are counting on the
% network to implicitly learn this mapping with low Lipschitz constant.
% %
% This is known to be a hard problem in the deep learning community~\citep{fazlyab2019efficient, khromov2023some, shi2022efficiently}.
% %
% Without enough training data\dataFootnote, models often end up in the overfitting regime (red)
% and generally have a large local Lipschitz constant, as shown in \cref{fig:method:regime}.
% %
% In fact, results in~\citep{shi2022efficiently} showed that neural networks (for
% classification) can have Lipschitz constants in the range of thousands despite appearing
% well-trained.
% %
% The problem is further complicated due to the point cloud-based nature of the network.
% %
% This goes against the intuition that the mapping $\phi \rightarrow \gamma_\phi^i$ should be smooth, and would
% limit the model's generalization ability~\citep{fazlyab2019efficient, khromov2023some}.
% %
% Meanwhile, our model's global Lipschitz constant is upper bounded
% by the Fourier coefficients $\mathbf{F}^i$, as stated in the following theorem:
%
Our representation of the concentration parameters $\gamma_\phi^i$ guarantees that 
each $\gamma_\phi^i$ is smooth with respect to $\phi$, as each $\gamma_\phi^i$ is 
a linear combination of sines and cosines.
Furthermore, the Fourier coefficients $\mathbf{F}^i$ provide an upper bound on
the Lipschitz constant $L^i$.
\begin{theorem}\label{thm:lipschitz}
    The concentration parameter $\gamma_\phi^i$ defined by the Fourier series
    representation in \cref{eqn:fourier} is Lipschitz continuous with respect to
    $\phi$ with an upper bound on the Lipschitz constant $L^i$ given by:
    \begin{equation}\label{eqn:lip_bound}
        L^i \leq \frac{1}{4} \sum_{k=1}^n k \sqrt{ (a_i^k)^2 + (b_i^k)^2 }.
    \end{equation}
\end{theorem}
\begin{proof}
    See \cref{sec:appendix:lipschitz}.
\end{proof}

\begin{wrapfigure}[6]{r}{0.25\textwidth}
    \centering
    \vspace{-1.8em}
    \includegraphics[width=0.75\linewidth]{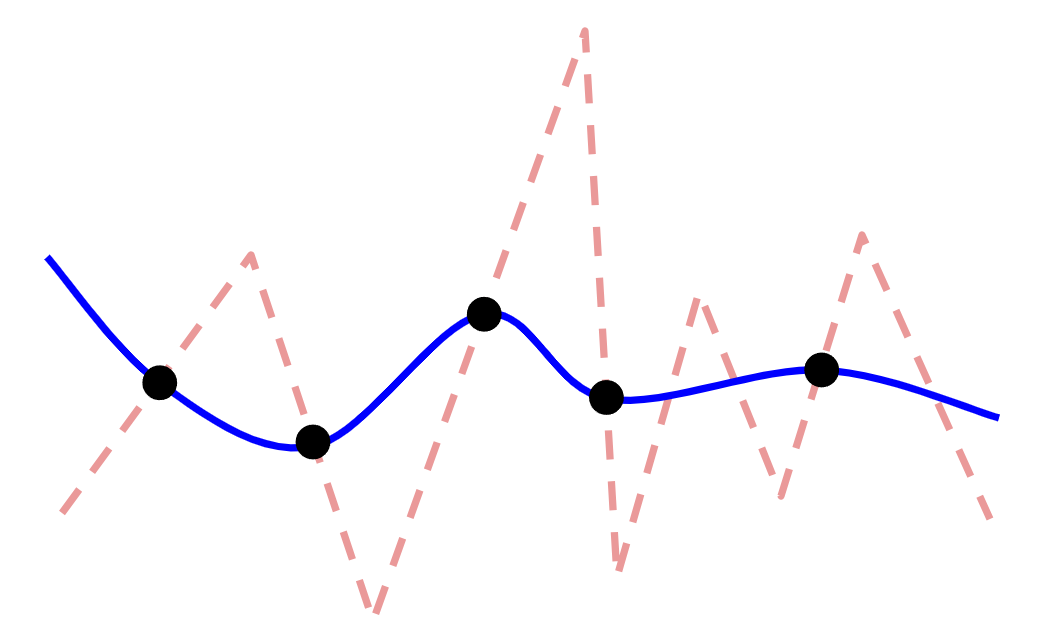}
    % \captionsetup{font=footnotesize}
    \vspace{-5pt}
    \caption{Demonstration of model in overfitting (red) and smooth-interpolation regime (blue).}
    \label{fig:method:regime}
\end{wrapfigure}
Having a bounded Lipschitz constant is desirable, as small Lipschitz constants
have been connected to improved generalization to unseen data in deep learning
\cite{belkin2019reconciling, fazlyab2019efficient, khromov2023some}.
With this
in mind, we consider \cref{thm:lipschitz} to suggest that our representation
improves the model's ability to generalize to unseen angles of approach $\phi$.
As SPARTA attempts to estimate risk based on both point cloud data and the angle of
approach $(\mathbf{Q}, \phi)$, we argue that an improved generalization over $\phi$
allows the learning process to focus on extracting meaningful information from the
point cloud $\mathbf{Q}$, rather than having to simultaneously learn the
relevant structure of the angle of approach $\phi$.
In other words, we believe that the smoothness and bounded Lipschitz constants
represent informative geometric priors that allows the model to more efficiently
learn relevant features.
We provide empirical evidence for this in \cref{sec:exp:performance}.

\subsection{Incorporating the Learned Risk Model into a Motion Planner}\label{sec:method:planner}
% \alanInline{I think this subsection is unfinished? Not going to give too much feedback here right now, but
%     let me know when to look at it. It is worth saying that we want to start this section by providing the high-level
%     motivation: why do we integrate our representation into a planner? What do we want to show?}
Let $\mathbf{x} \in \mathcal{X} \subseteq \mathbb{R}^s$ denote the state of the robot,
$\mathbf{u} \in \mathcal{U} \subseteq \mathbb{R}^u$ the control input,
and subscript $t \geq 0$ the timestep.
We assume the robot's motion evolves as a discrete time system, $\mathbf{x}_{t+1} = F(\mathbf{x}_t, \mathbf{u}_t)$.
% \begin{equation}
%     \mathbf{x}_{t+1} = F(\mathbf{x}_t, \mathbf{u}_t).
% \end{equation}
%
The planner optimizes for a control input sequence $\mathbf{u}_{0:T}$ given an initial state $\mathbf{x}_0$ that minimizes
the worst-case planning objective.
Specifically, we break the cost function into two parts:
A task-related term $C(\mathbf{x}_{0:T})$ (e.g., goal reaching cost) and the worst-case
risk computed from the risk distribution $\mathbf{p}(\gamma_{\phi_t})$.
We adopt the Conditional Value at Risk (CVaR) as the worst-case risk because it is coherent
and thus suitable for trustworthy risk assessment in robotics~\citep{majumdar2020should}.
Specifically, the CVaR at level $\alpha \in [0, 1]$ of the risk variable $\gamma_\phi$
is $\text{CVaR}_\alpha(\gamma_\phi) := \frac{1}{\alpha} \int_0^\alpha \min\{\gamma \mid p(\gamma_\phi > \gamma) \leq \tau \} d\tau$,
% \begin{align}\label{eqn:var}
%     \text{CVaR}_\alpha(\gamma_\phi) & := \frac{1}{\alpha} \int_0^\alpha \min\{\gamma \mid p(\gamma_\phi > \gamma) \leq \tau \} d\tau,
% \end{align}
%
where the probabilities can be estimated from $\mathbf{p}(\gamma_\phi)$.
Intuitively, $\text{CVaR}_\alpha(\gamma_\phi)$ represents the expectation of the
last (right tail) $\alpha$ portion of the distribution $\mathbf{p}(\gamma_\phi)$.
Notice that while the categorical distribution is discrete, so its CVaR is not necessarily
continuous w.r.t.~$\phi$, our formulation guarantees the CVaR does not suffer from a wrap-around 
discontinuity, as shown in \cref{fig:method:wrap-around:b}.
To summarize, we adopt MPPI~\citep{williams2016aggressive,williams2017information} with the following planning problem:
\begin{equation}
    \begin{alignedat}{3}\label{eqn:planner}
        \min_{\mathbf{u}_{0:T}} \quad & C(\mathbf{x}_{0:T}) + \sum_{t=0}^T \text{CVaR}_\alpha(\gamma_{\phi_t})                                                                 \\
        \text{s.t.} \quad
                                      & \mathbf{x}_{t+1} = F(\mathbf{x}_t, \mathbf{u}_t)                                  &  & \quad \quad \text{(Dynamics)}                   \\
                                      & \mathbf{F}^i_{t+1} = \Gamma_\theta(\mathbf{Q}_{t+1})[i], ~ \forall ~ i \in [1, B] &  & \quad \quad \text{(Fourier Coefficients)}       \\
                                      & \gamma_{\phi_{t+1}}^i \gets \cref{eqn:fourier}, ~ \forall ~ i \in [1, B]          &  & \quad \quad \text{(Risk Concentration Params.)} \\
                                      & \mathbf{p}_{t+1}(\gamma_{\phi_{t+1}}) \gets \cref{eqn:PMF}                        &  & \quad \quad \text{(Risk PMF)}
    \end{alignedat}
\end{equation}

\section{Experiments}\label{sec:exp}
In this section, we demonstrate the application of the proposed framework to assess potential vehicle damage 
(e.g., flat tires, broken axle).
Although vehicle damage frequently occurs during real off-road autonomous operations, previous research has largely overlooked this risk metric or relied on relatively simple handcrafted heuristics (such as stepwise elevation change) to model it~\citep{fan2021step, datar2024learning, lee2025trg}.

\subsection{Tire Deformation As A Surrogate Metric}\label{sec:exp:data_collection}
% \alanInline{
%     Currently, this reads as if we are choosing a metric (tire deformation)
%     because it happens to work well for our methodology (i.e., it has obvious
%     bounds). This implies that what we can do is limited by our methodology.
%     Our writing should instead say that we chose a certain metric for good
%     practical reasons and that we should be able to handle many other useful
%     metrics as well.
% }
In the following experiments, we use tire deformation extent as a surrogate metric for the probability
of vehicle damage.
While our formulation can handle other risk-related variables (\cref{sec:appendix:example}) like wheel force and acceleration,
which are also related to terrain contact, we use tire deformation because the tire is the part
that directly contacts the terrain and thus is less susceptible to noise accumulated from simulating
other vehicle parts.
We collect training data using a high-fidelity simulation platform BeamNG.tech\footnote{We choose BeamNG over
conventional simulators for its high-fidelity soft body physics simulation.}~\citep{beamng_tech}.
Specifically, for a tire node $w$, let $r_w$ denotes the distance from the (deformed) tire node 
to wheel center, $r_\text{inner}$ and $r_\text{outer}$ the radius of the wheel frame and non-deformed 
tire, respectively, and we compute the deformation extent as $d_w = \frac{r_w - r_\text{inner}}{r_\text{outer} - r_\text{inner}}$.
% \begin{equation}\label{eqn:deformation}
%     d_w = \frac{r_w - r_\text{inner}}{r_\text{outer} - r_\text{inner}},
% \end{equation}
%
\cref{sec:appendix:data} includes further details of the data collection process.

\subsection{Model Performance}\label{sec:exp:performance}
% \textbf{Rework this after writing Lipschitz Continuous property, should connect to that theorem.}

% \alanInline{I really like the way that you open this section by stating what the goal is.
%     I would take it even further! Have a short single sentence that says the
%     goal of this section. You can then follow it up with a sentence that
%     tells the reader \emph{how} you try to achieve this goal.}

\begin{wrapfigure}[10]{r}{0.25\textwidth}
    \centering
    \vspace{-1.2em}
    \includegraphics[width=0.85\linewidth]{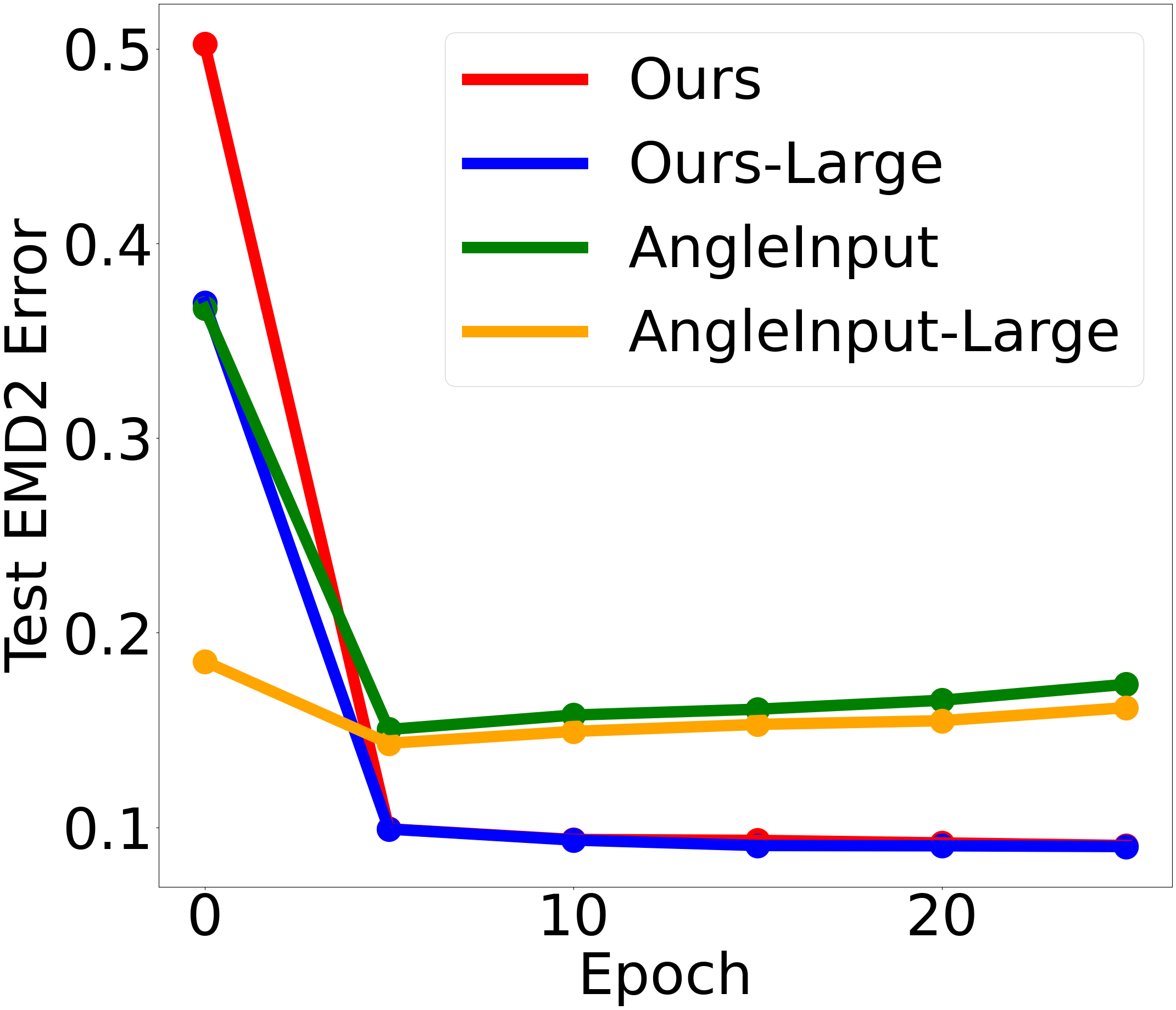}
    \caption{
        Our Model (Red and Blue) achieve lower test $\text{EMD}^2$ loss compared to \textit{AngleInput} (Green and Orange) without overfitting. 
    }
    \label{fig:exp:EMD2}
\end{wrapfigure}

To demonstrate the generalization capability of our method,
we compare its test set performance against a model that encodes the angle of approach as an input
(\textit{AngleInput}), where our model outperforms \textit{AngleInput}.
Our model and the \textit{AngleInput} share the same architecture and have approximately the
same number of trainable parameters, except that the \textit{AngleInput} has an additional encoder
head for the angle of approach, which is fused downstream with the point cloud feature
to predict the concentration parameters.
Details of the model architecture can be found in \cref{sec:appendix:model}.
In \cref{fig:exp:EMD2}, we show the test $\text{EMD}^2$ loss achieved by our model and the \textit{AngleInput}
with varying model sizes.
The \textit{AngleInput} baselines (Green and Orange) quickly overfit to the training data, with a best test loss of 0.15.
In contrast, our test loss (Red and Blue) continues to decrease and eventually converged to 0.09.
This is aligned with our hypothesis in \cref{sec:method:theory}, that our model would generalize better
compared to \textit{AngleInput} because the Fourier basis functions promise an upper-bound for 
the Lipschitz constant of the mapping $\phi \rightarrow \gamma_\phi^i$, whereas the \textit{AngleInput} has to
learn this mapping implicitly.
\par

\subsection{Planning in Simulation}\label{sec:exp:sim}
\begin{wrapfigure}[9]{r}{0.38\textwidth}
    \centering
    % \vspace{-1.2em}
    \begin{subfigure}[t]{0.2\textwidth}
        \centering
        \includegraphics[width=\linewidth]{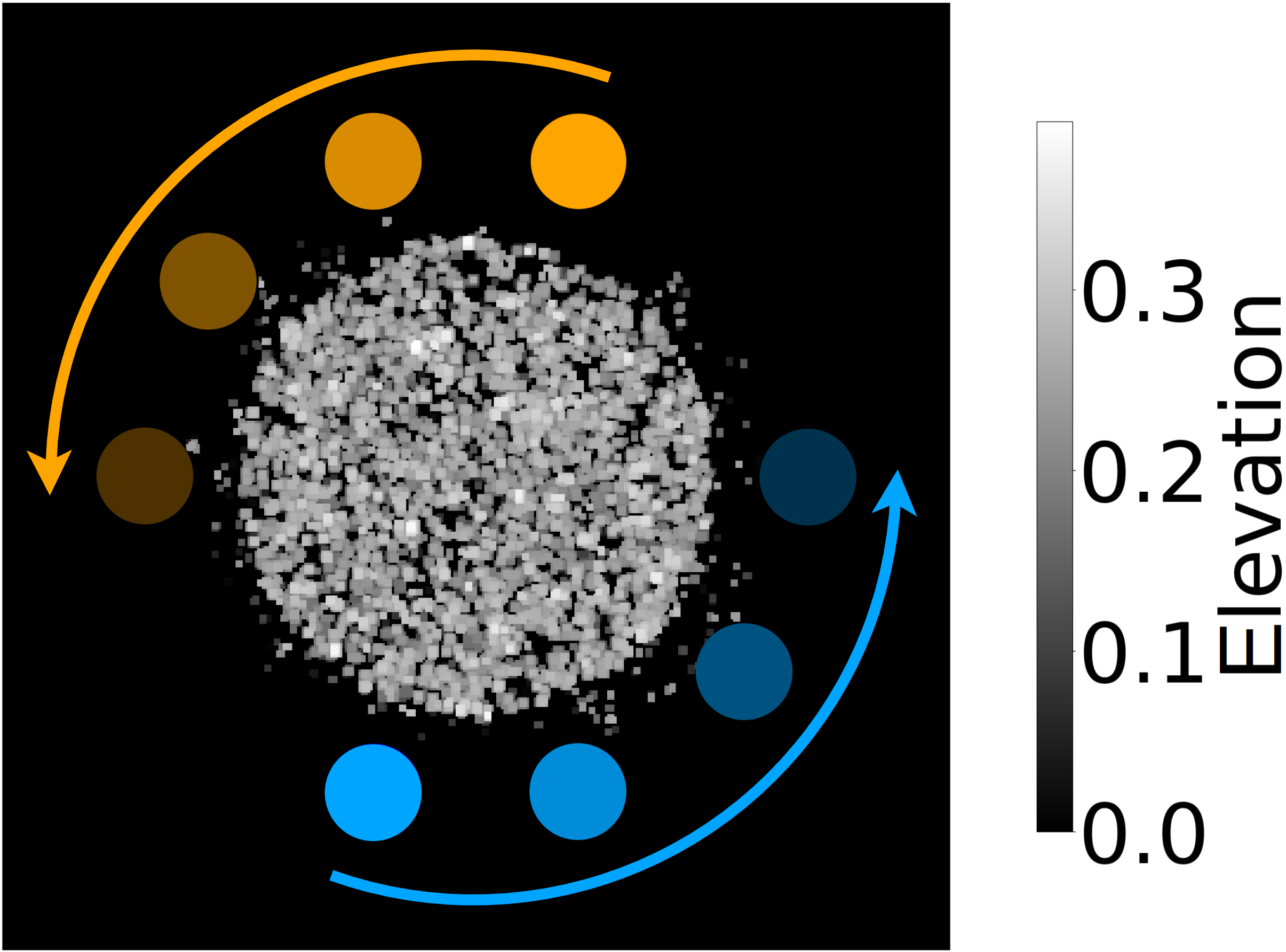}
        \captionsetup{skip=0pt}
        \caption{}
        \label{fig:exp:boulder_field:elev}
    \end{subfigure}
    \hfill
    \begin{subfigure}[t]{0.15\textwidth}
        \centering
        \includegraphics[width=\linewidth]{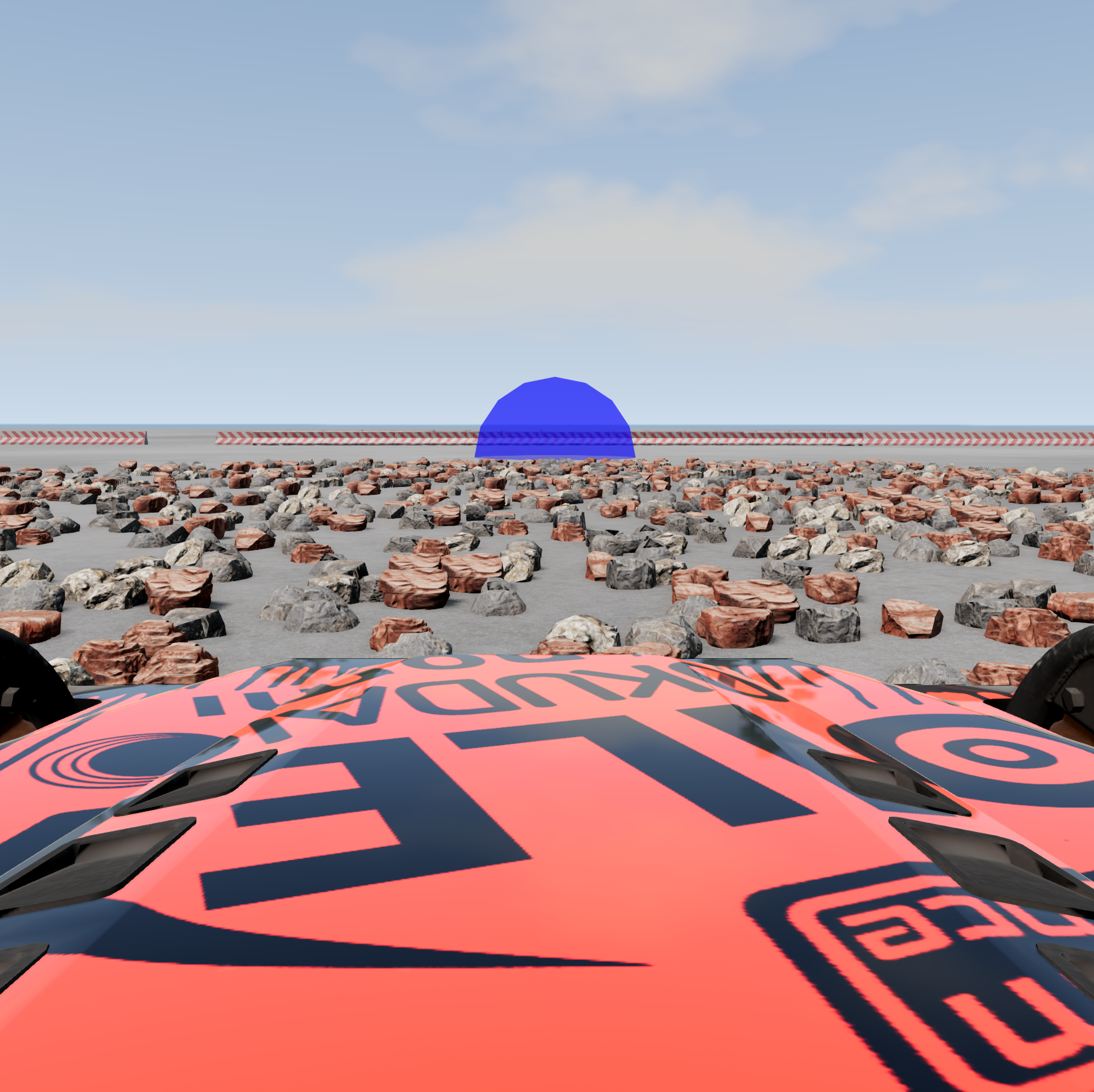}
        \captionsetup{skip=0pt}
        \caption{}
        \label{fig:exp:boulder_field:view}
    \end{subfigure}
    \vspace{-5pt}
    \caption{(a): Overview of the Boulder Field (40m) test environment with goals (blue) and starting points (orange),
    overlaid with an elevation map (grayscale). (b): Overview of the environment from driver's POV.}
    \label{fig:exp:boulder_field}
\end{wrapfigure}

% \begin{wrapfigure}[9]{r}{0.38\textwidth}
%     \centering
%     % \vspace{-1.2em}
%     \begin{subfigure}[t]{0.2\textwidth}
%         \centering
%         \includegraphics[width=\linewidth]{images/experiment/boulder_field_env_matching_ratio.png}
%         \captionsetup{skip=0pt}
%         \caption{}
%         \label{fig:exp:boulder_field:elev}
%     \end{subfigure}
%     \hfill
%     \begin{subfigure}[t]{0.15\textwidth}
%         \centering
%         \includegraphics[width=\linewidth]{images/experiment/driver_view_new.png}
%         \captionsetup{skip=0pt}
%         \caption{}
%         \label{fig:exp:boulder_field:view}
%     \end{subfigure}
%     % \includegraphics[width=\linewidth]{images/experiment/boulder_field_env_cap.png}
%     \vspace{-5pt}
%     \caption{(a): Overview of the Boulder Field (40m) test environment with goals (blue) and starting points (orange),
%     overlaid with an elevation map (grayscale). (b): Overview of the environment from driver's POV.}
%     \label{fig:exp:boulder_field}
% \end{wrapfigure}
To test integration with a planner, we designed a challenging task where the vehicle
needs to cross a randomly generated boulder field, with 20m radius and 1500 randomly placed
obstacles, from 100 uniformly placed starting positions.
%
% To test integration with a planner, we designed a challenging task where the vehicle 
% needs to cross a randomly generated boulder field to reach its destination.
% %
% The boulder field is a circular region filled with 1,500 random obstacles, with 100 
% starting and goal positions placed uniformly around the circle.
%
%
\cref{fig:exp:boulder_field} shows the overview of the environment.
The obstacles are placed so densely that finding a clear path is challenging. 
Therefore, to complete the task at high speed, the planner must accurately distinguish between obstacles that 
can be safely traversed and those that pose significant risks.
In \cref{sec:appendix:open_loop} we provide an open loop experiment with ablation study to further demonstrate
our approach.
Besides \textit{AngleInput}, we consider 2 baselines: (i) a model trained to predict the 
distributions without modeling the angle dependency (\textit{AngleFree}), and (ii) a heuristic baseline that 
chooses the point with lowest maximum elevation under vehicle's wheel footprint (\textit{Elev}).
Since the obstacles are placed on flat ground, \textit{Elev} is equivalent to evaluating stepping difficulty as in~\citep{fan2021step, lee2025trg}.
We use a MPPI~\cite{williams2016aggressive} planner with Ackermann kinematics, and the planning objective includes the distance 
to the goal, a velocity cost for driving over speed limit $\bar{v}$, and a risk cost:
\begin{align}
    C_\text{risk}(t) & = \left\{ \begin{array}{rcl}
        & \text{CVaR}_\alpha(\gamma_{\phi_t}) & \mbox{for Learned Models} \\
        & \max(h_\text{FL}, h_\text{FR}, h_\text{RL}, h_\text{RR}) & \mbox{for Elevation Baseline}
    \end{array}\right. \\
    Cost & = \underbrace{w_\text{goal}\sum_{t=0}^T D(\mathbf{x}, \mathbf{x}_\text{goal})}_{\text{Goal Cost}} + \underbrace{w_\text{v} \sum_{t=1}^T \mathbb{1}(v_t > \bar{v})}_{\text{Velocity Cost}} + \underbrace{w_\text{risk} \sum_{t=0}^T v_t C_\text{risk}(t)}_{\text{Risk Cost}}. \label{eqn:objective_sim}
\end{align}
where $w_\text{goal}, w_\text{v}, w_\text{risk}$ are weights for associated cost term. 
% %
% We include the velocity cost because we use throttle and steering angle as control 
% input due to limitation of the simulator, as opposed to linear and angular velocity.
%
Notice we scale the Risk Cost term by the current vehicle velocity $v_t$ to penalize traversing risky terrain at high speed.

\begin{wraptable}[10]{r}{0.45\textwidth}
    \centering
    \vspace{-1.2em}
    \footnotesize
    \begin{tabular}{c | c | c c}
        \hline
        Algorithm   & $\bar{v} $ & S.R. (\%) $\uparrow$ & Vel (m/s) $\uparrow$ \\
        \hline
        Ours                           &  4  &  \textbf{91}    &   \underline{2.6}     \\
        \hline
        AngleInput                            &  4  &  \underline{85} &   2.5                 \\
        \hline
        \multirow{2}{4em}{AngleFree}  &  4  &  73             &   \textbf{3.2}        \\
                                       &  2  &  79             &   1.9                 \\
        \hline
        \multirow{2}{2em}{Elev}        &  4  &  73             &   \textbf{3.2}        \\
                                       &  2  &  75             &   2.1                 \\
        \hline
    \end{tabular}
    \caption{Success rate (S.R.) of the models in the boulder field test.
    Our method achieves best success rate, and overall angle-dependent models perform better than those that do not consider angle. 
    \textbf{Best}, and \underline{second best}.}
    \label{tab:exp:boulder_field}
\end{wraptable}
In \cref{tab:exp:boulder_field} we report the success rate (S.R., reached goal without damage) and average vehicle velocity (Vel)
for the methods tested. 
Our model achieves higher success rate compared to \textit{AngleInput}, presumably because of its stronger generalization ability.
Most importantly, the performances of models that captures angle-dependency (ours and \textit{AngleInput}) are
better than \textit{AngleFree} and \textit{Elev} because their Risk Cost term cannot distinguish between safe obstacles from risky ones.
As a result, the vehicle attempts to traverse the boulder field at a roughly constant velocity, i.e. the Risk Cost 
in these cases degenerates to velocity cost.
To test if the faster speed of \textit{AngleFree} and \textit{Elev} caused their low success rates, we
tuned the weights and decreased $\bar{v}$ from 4 to 2 to enforce a slower velocity.
However, their success rates do not significantly increase even traversing at a lower velocity, 
which suggests that they are ineffective risk estimators in this environment.

\begin{figure}[t]
    \label{fig:exp:crafted}
    \centering
    \vspace{-1.0em}
    \begin{subfigure}[t]{0.35\linewidth}
        \centering
        \includegraphics[width=\linewidth]{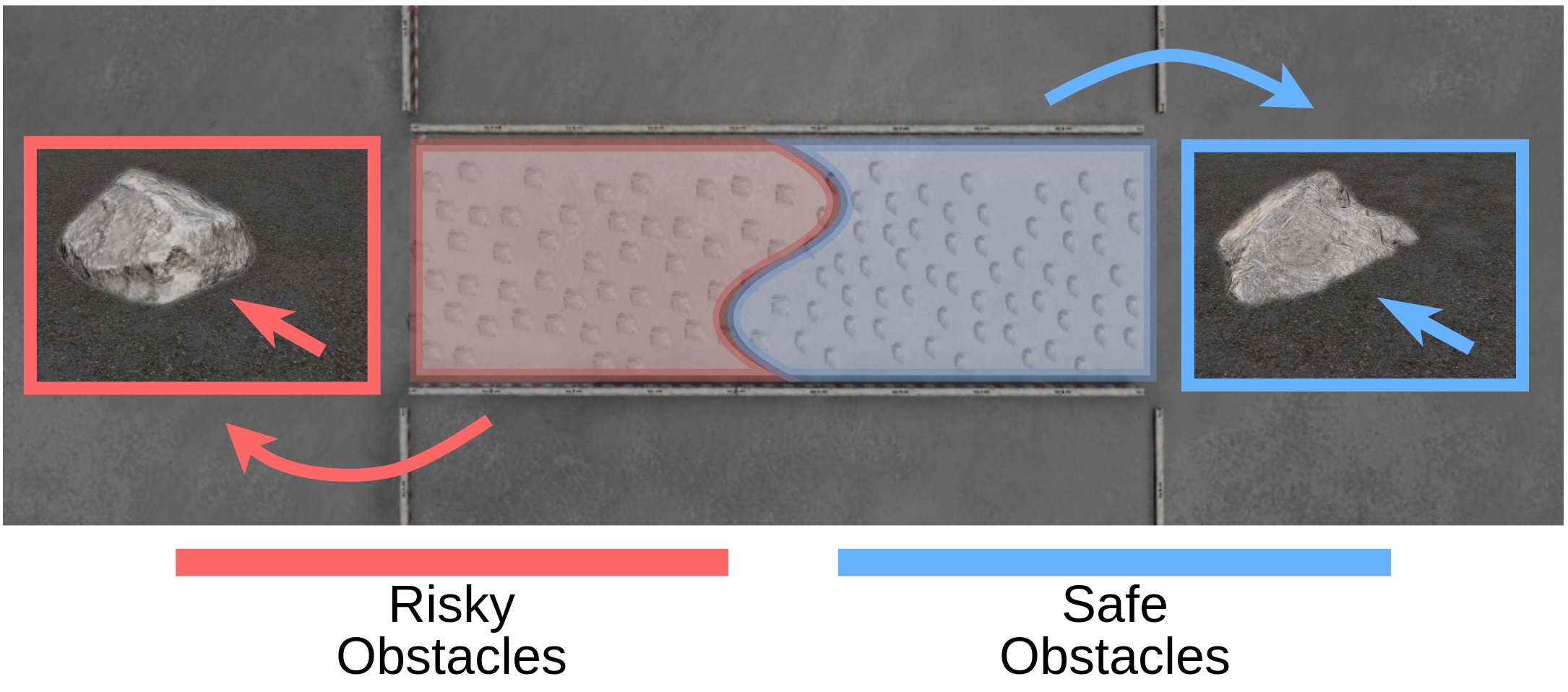}
        \captionsetup{skip=0pt}
        \caption{}
        \label{fig:exp:crafted_env}
    \end{subfigure}
    \hspace{0.04\linewidth}
    \begin{subfigure}[t]{0.35\linewidth}
        \centering
        \includegraphics[width=\linewidth]{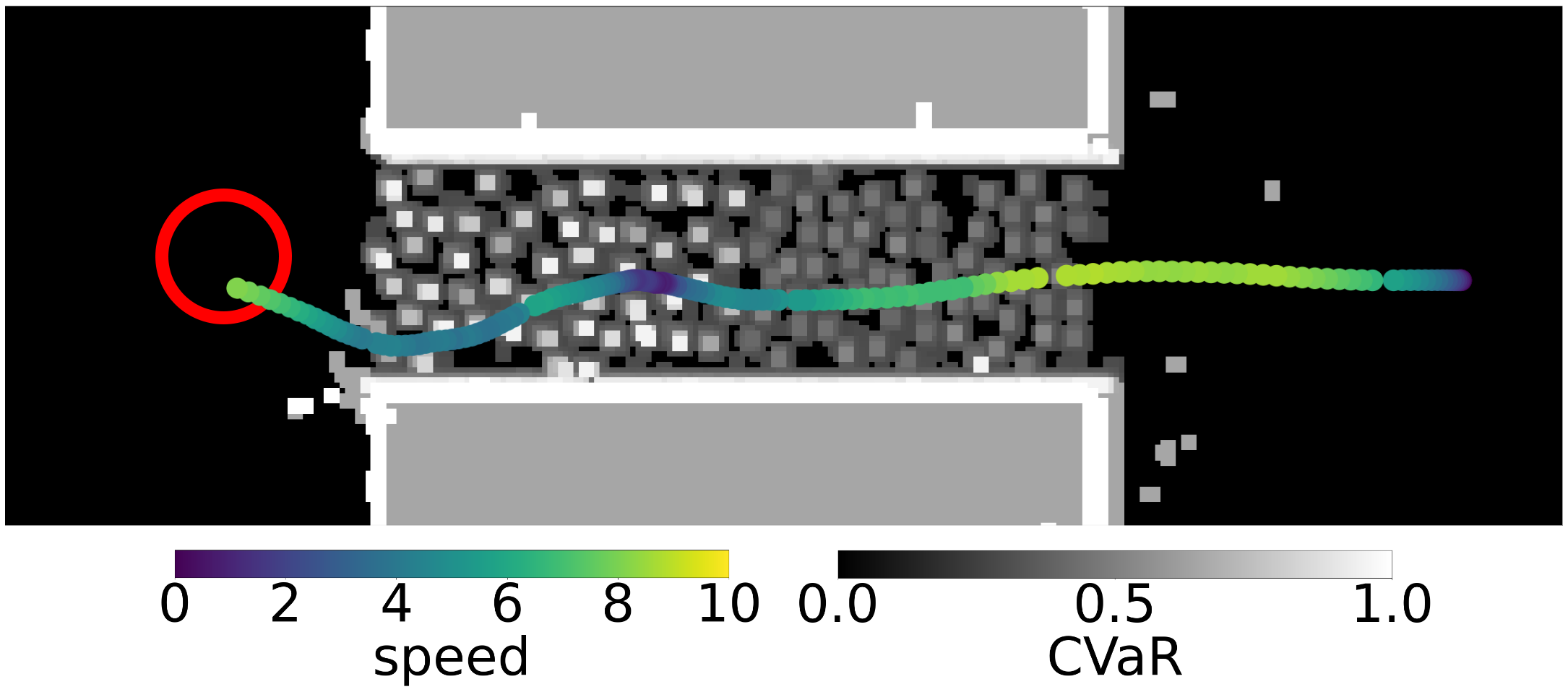}
        \captionsetup{skip=0pt}
        \caption{}
        \label{fig:exp:crafted_ours}
    \end{subfigure}
    \begin{subfigure}[t]{0.35\linewidth}
        \centering
        \includegraphics[width=\linewidth]{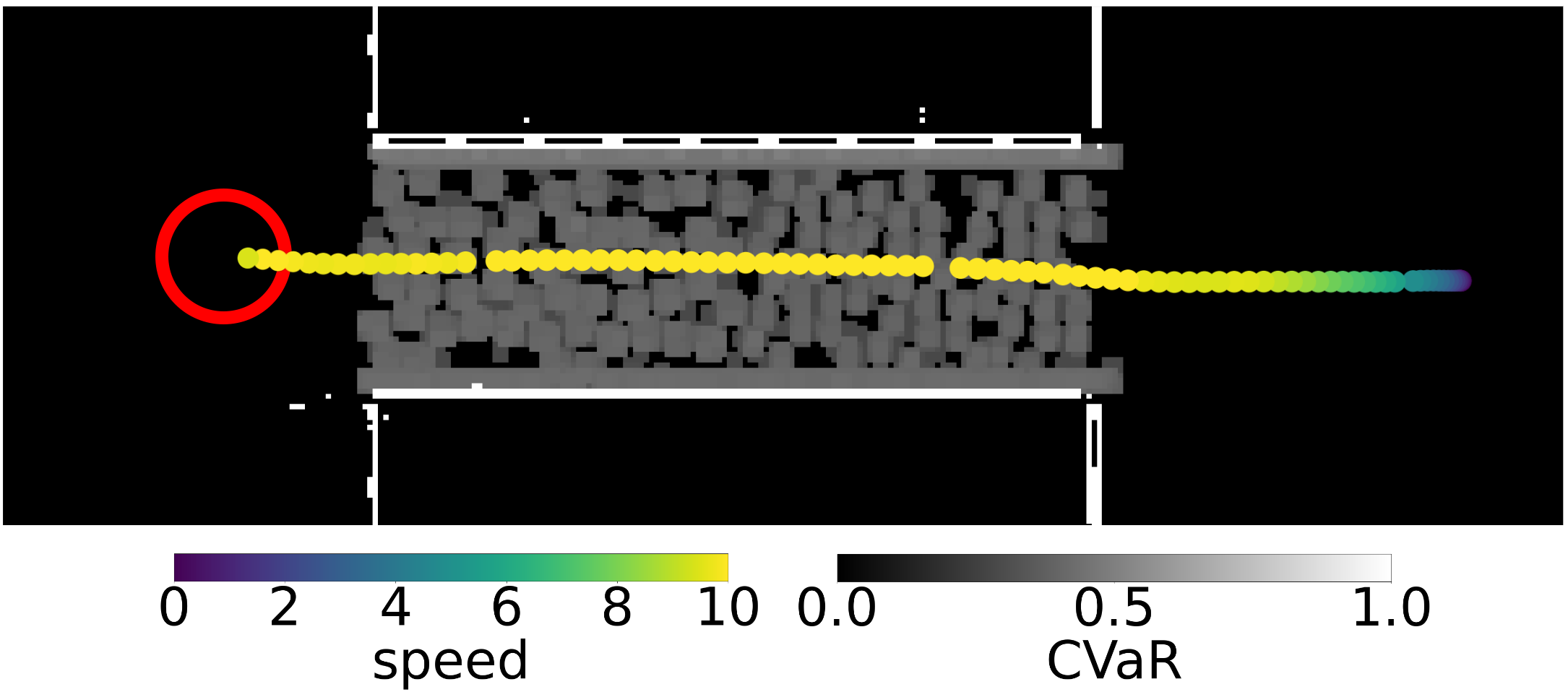}
        \captionsetup{skip=0pt}
        \caption{}
        \label{fig:exp:crafted_angle_free}
    \end{subfigure}
    \hspace{0.04\linewidth}
    \begin{subfigure}[t]{0.35\linewidth}
        \centering
        \includegraphics[width=\linewidth]{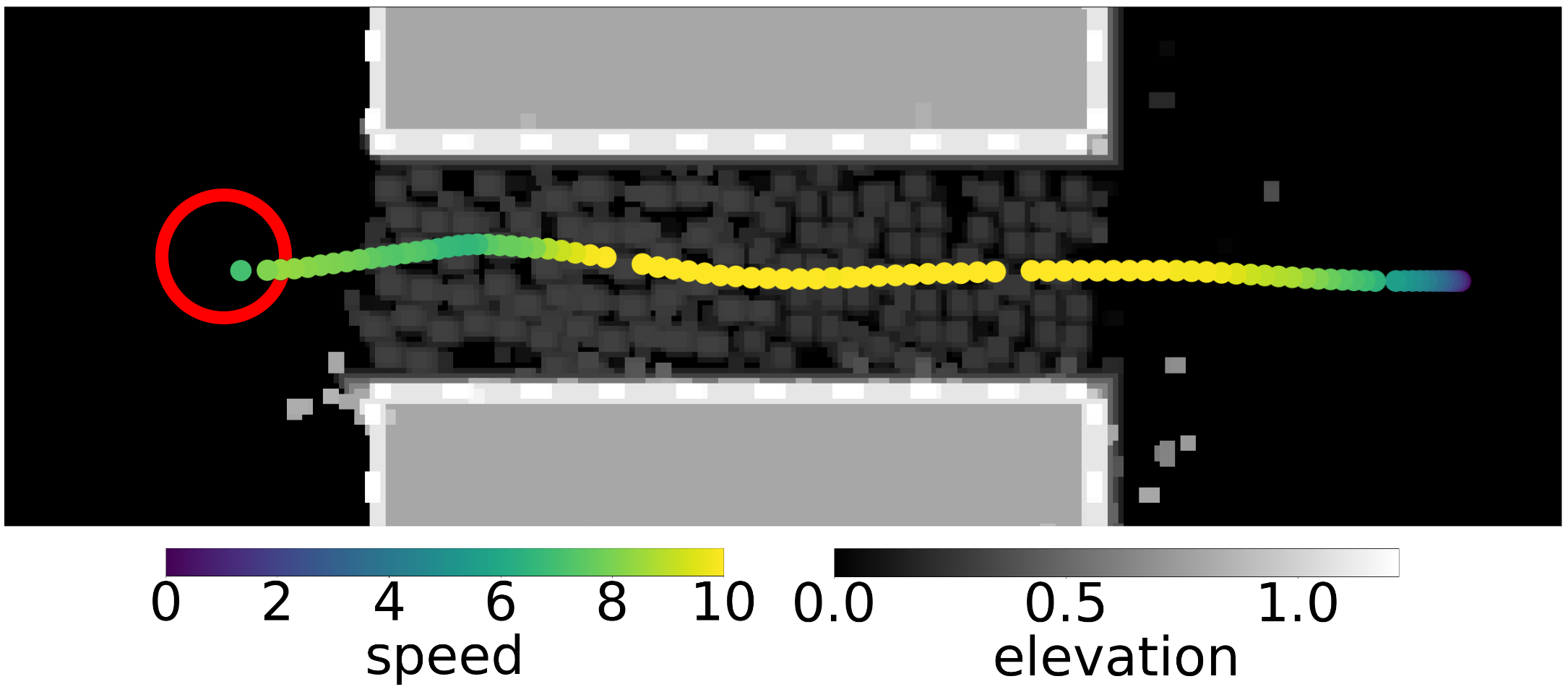}
        \captionsetup{skip=0pt}
        \caption{}
        \label{fig:exp:crafted_elev}
    \end{subfigure}    
    \vspace{-5pt}
    \caption{
        (a): Overview of the environment. The left half of the environment is filled with risky obstacles (Red),
        and the right half is filled with safe ones (Blue). Arrow indicates expected angle of approach. 
        (b): Background denotes CVaR predicted by our method at the expected angle of approach. The planner is
        able to identify risky obstacles and decelerate preemptively. \textbf{Max Tire Force: 1426 N}.
        (c): Background denotes CVaR predicted by the \textit{AngleFree}. It can not
        identify risky obstacles and the vehicle drives at a constant speed. \textbf{Max Tire Force: 2937 N}.
        (d): Background denotes elevation. Elevation baseline treat them as equally risky and traverse 
        with a constant speed. \textbf{Max Tire Force: 4367 N}.
    }
    \vspace{-1.8em}
\end{figure}
To further help understand how \textit{AngleFree} and \textit{Elev} underestimate traversal risk, we crafted an
environment, where the vehicle has to go through a narrow passage way filled with obstacles.
We picked two obstacles with same elevation yet one appears to be safe and the other is visually risky,
and use each of them fill half of the environment.
Detail of the environment is shown in \cref{fig:exp:crafted_env}, where red denotes the risky obstacles and blue denotes
safe obstacles, when approached from the expected orientation (from right to left).
The CVaR predicted by our model (\cref{fig:exp:crafted_ours}) correctly identifies the risky obstacles from the safe ones,
and is manifested by the vehicle's preemptive deceleration.
During the traversal, our model induced the lowest force on the tire 1426 N.
The CVaR predicted by the \textit{AngleFree} model (\cref{fig:exp:crafted_angle_free}) however, is very similar for all obstacles
because the obstacles are all safe to traverse from certain orientations,
and as a result the vehicle drives at a constant (high) velocity through the obstacles.
Similar to \textit{AngleFree}, the obstacles have same height and thus \textit{Elev} drives the vehicle at a 
constant speed (until it hits an obstacle very hard and thus slowed down).
Both \textit{AngleFree} and \textit{Elev} results in significantly higher tire force (2937 N and 4367 N respectively),
signifying higher probability of vehicle damage.
In other words, in order for \textit{AngleFree} and \textit{Elev} to safely drive the vehicle
through complex obstacles, their maximum speed must be set conservatively such that the vehicle can
safely traverse the most risky obstacle possible.
However, empirically it is difficult to find such a threshold that guarantees safety.

\subsection{Planning on Hardware}
\begin{wrapfigure}[15]{r}{0.5\textwidth}
    \centering
    \vspace{-1.2em}
    \includegraphics[width=0.9\linewidth]{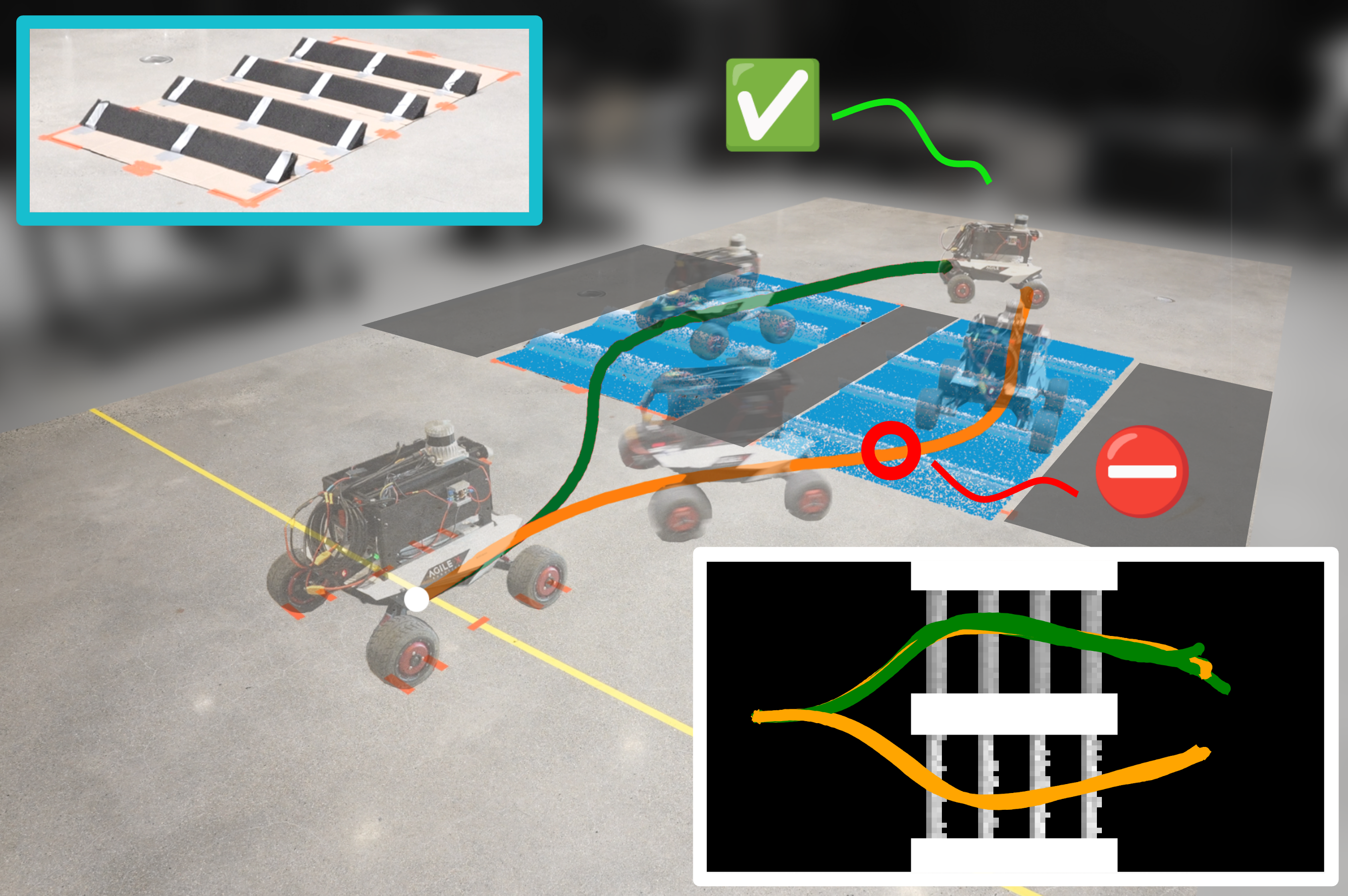}
    \caption{Environment overview with overlaid point cloud (blue, colored by elevation), physical obstacle (cyan box),
    and lethal obstacles (shaded). Our method (Green) consistently chooses the safe path, but \textit{Elev} (Orange) 
    can choose the risky path as the obstacles have same elevation change. Trajectories
    are shown in the white box and background denotes CVaR of the expected angle of traversal (Left to Right).}
    \label{fig:exp:traj_in_high_bay}
\end{wrapfigure}
In this experiment, we demonstrate our model on a custom-built wheeled robot (specification can be found in \cref{sec:appendix:hardware}), and
compare its performance qualitatively to \textit{Elev}.
We create an environment where a robot has to pick one out of 2 possible paths to reach its goal, where the left path is safe and
the right path is risky.
For the obstacle, we cut rigid foam into a shape resembling a traffic flow plate. 
We place the obstacle on the left path as shown in the cyan box in \cref{fig:exp:traj_in_high_bay}.
To avoid vehicle damage and allow the robot to smoothly traverse, we \emph{virtually} place the same obstacle rotated 180 degrees on the 
right path as the risky obstacle.
The point clouds (pre-collected with~\citep{chen2023dlio}) are overlaid in \cref{fig:exp:traj_in_high_bay} for demonstration purposes.
The planner uses unicycle dynamics and the planning objective is the same as \cref{eqn:objective_sim} except we do not include
the velocity terms.
The planner samples 500 rollouts with a horizon of 5.0s, and runs at 50 Hz with our model (70 Hz with \textit{Elev}).
In the white box in \cref{fig:exp:traj_in_high_bay} we show the trajectories of our method (Green) and \textit{Elev} (Orange).
Our model correctly identifies the safe path for all 5 trials, yet \textit{Elev} chooses the risky path for 3 out of 5 trials.
While our model drives the vehicle safely and smoothly through the safe obstacles, \textit{Elev} would 
have likely caused vehicle damage by ramming into sharp edges of an obstacle (red circle).

% \begin{wrapfigure}[13]{r}{0.35\textwidth}
%     \centering
%     \vspace{-2.4em}
%     \includegraphics[width=\linewidth]{images/experiment/2d_traj.png}
%     \caption{White region denotes (virtual) lethal obstacles, and bright background denotes higher CVaR at 
%     expected angle of approach (arrow). Our model (green) consistently chooses the safe path but \textit{Elev} (orange) chooses randomly
%     as the obstacles have same elevation.}
%     \label{fig:exp:2d_traj}
% \end{wrapfigure}

% %===============================================================================

\section{Conclusion}
In this paper, we propose \textbf{SPARTA}, a general framework for estimating terrain traversal risk
conditioned on vehicle angle of approach.
Specifically, we identify and exploit the geometric structure in the problem setting, and propose
to predict a smooth analytical function over the 1-Sphere to achieve better data efficiency and generalization
performance.
During planning, the function is queried to construct a categorical distribution for a risk variable of interest at
any angle of approach, and the distribution's tail risk (CVaR) is considered to account for aleatoric uncertainty.
Experiments in simulation and on hardware demonstrate the effectiveness of the proposed approach.
Most importantly, off-road problems are naturally rich in geometric structure, and our formulation is general enough
such that it finds application in numerous different problem settings.
Future work may further explore the geometric structure in other off-road autonomy problems.

% %===============================================================================

\clearpage
\section{Limitation}
Our work has several limitations.
For example, in this work we assume the environment is fully observable, i.e. a prebuilt point cloud
map is accessible during planning.
While the point cloud map we can build on the fly is sufficiently dense, the model's performance without
a prebuilt map is still to be tested.
A potential solution to this problem is to penalize terrain with limited observation during planning or
incorporate depth inpainting as in~\citep{frey2024roadrunner, meng2023terrainnet}.
In addition, this work focused on estimating traversability given geometric information, whereas semantic information
is also important to consider in many off-road settings.
We anticipate that future work could incorporate semantic information by performing sensor fusion with point pillar feature
in birds-eye-view~\citep{frey2024roadrunner}.
While the smoothness assumption in \cref{sec:method:theory} is reasonable, settings that potentially break
this assumption might exist in the real world. Quantitative analysis to identify such failure modes can be a valuable 
addition to our framework.
Last but not least, tire deformation is a metric that's only available in simulation.
However, we are not interested in predicting tire deformation in the real world, but rather using it as a 
surrogate metric for terrain property (i.e. how risky is the terrain).
Future work could explore other variables to quantify vehicle damage.

% %===============================================================================

% \section{Citations}
% \label{sec:citations}

% 	Citations can be made using either \textbackslash citep\{\} or \textbackslash citet\{\}, depending from the appropriateness. To avoid the citation moving to the next line, it is often a good practice to replace the space before with a tilde (\~{}) character.
% 	Example 1: ``CoRL is the best conference ever~\citep{oliveira2021advances}.''
% 	Example 2: ``\citet{oliveira2021advances} proved, both theoretically and numerically, that CoRL is the best conference ever.''

% %===============================================================================

% \section{Experimental Results}
% \label{sec:result}

% %===============================================================================

% \section{Conclusion}
% \label{sec:conclusion}

% %===============================================================================

% The acknowledgments are automatically included only in the final and preprint versions of the paper.
\acknowledgments{This research was sponsored by the DEVCOM Army Research Laboratory (ARL) under SARA CRA
W911NF-24-2-0017. Distribution Statement A: Approved for public release; distribution is unlimited.}

\bibliography{reference}

%===============================================================================

\clearpage
% \documentclass{article}
% \usepackage{amssymb}

% \usepackage{corl_2025} % Use this for the initial submission.
% % \usepackage[final]{corl_2025} % Uncomment for the camera-ready ``final'' version.
% %\usepackage[preprint]{corl_2025} % Uncomment for pre-prints (e.g., arxiv); This is like ``final'', but will remove the CORL footnote.

% \usepackage{amsmath,amssymb,amsfonts, amsthm}
% \usepackage{graphicx}
% \usepackage{wrapfig}
% \usepackage[font=scriptsize,labelsep=period]{caption}
% \usepackage[font=scriptsize,labelsep=period]{subcaption}
% \usepackage{subfiles}
% \usepackage[capitalize]{cleveref}
% \usepackage{multicol}
% \usepackage{multirow}
% \usepackage{bbold}

% \begin{document}

\appendix

\section{Derivation of Lipschitz Constant} \label{sec:appendix:lipschitz}
With a slight abuse of notation $F$, \cref{eqn:fourier} can be written in the following way:
\begin{align}
    G(\phi) & = \frac{a_i^0}{2} + \sum_{k=1}^n [a_i^k \cos(k\phi) +b_i^k \sin(k\phi) ] \\
    F(\phi) & = \sigma(G(\phi))
\end{align}
where $\sigma(\cdot)$ is the sigmoid function and $a_i^0$,
$a_i^k$, and $b_i^k$ are the Fourier coefficients predicted by our network.

We will derive an upper bound on the Lipschitz constant of $F(\phi)$, via the chain rule:
\begin{align}
    |F^\prime(\phi)| & = |\sigma'(G(\phi))| \cdot |G'(\phi)|.
\end{align}

First we differentiate $G(\phi)$ w.r.t. $\phi$:

\begin{align}
    G'(\phi) & = \sum_{k=1}^n ka_i^k\cos(k\phi) - k b_i^k \sin(k\phi) \\
             & = \sum_{k=1}^n k [a_i^k\cos(k\phi) - b_i^k \sin(k\phi)]. \label{eqn:F_prime}
\end{align}

For every $k \in [1, n]$, let:

\def\phaseOffset{\Delta_k}
\begin{align}
    R_k & = \sqrt{ (a_i^k)^2 + (b_i^k)^2 } \\
    \phaseOffset & = \arctan(\frac{-b_i^k}{a_i^k}),
\end{align}

Then \cref{eqn:F_prime} can be rewritten as a phase shift leveraging the cosine addition formula:

\begin{align}
    G'(\phi) & = \sum_{k=1}^n k [ R_k \cos(k\phi) \frac{a_i^k}{R_k} - R_k \sin(k\phi) \frac{b_i^k}{R_k}] \\
             & = \sum_{k=1}^n k [ R_k \cos(k\phi) \cos(\phaseOffset) + R_k \sin(k\phi) \sin(\phaseOffset)] \\
             & = \sum_{k=1}^n k R_k \cos(k\phi - \phaseOffset). \label{eqn:phase_shift}
\end{align}

Notice that the maximum value of a $\cos(\cdot)$ is 1, as a result, \cref{eqn:phase_shift} is upper bounded by:

\begin{align}
    G'(\phi) & \leq \sum_{k=1}^n k R_k \\
             & = \sum_{k=1}^n k \sqrt{ (a_i^k)^2 + (b_i^k)^2 }.
\end{align}

Because the sigmoid function is Lipschitz with a Lipschitz constant of $\frac{1}{4}$, by the chain rule:

\begin{align}
    |F^\prime(\phi)| & = |\sigma'(G(\phi))| \cdot |G'(\phi)| \\
                & \leq \frac{1}{4}  \sum_{k=1}^n k \sqrt{ (a_i^k)^2 + (b_i^k)^2 }.
\end{align}

As a result the Lipschitz Constant of $F(\phi)$ is bounded by:

\begin{equation}
    L^i \leq \frac{1}{4}  \sum_{k=1}^n k \sqrt{ (a_i^k)^2 + (b_i^k)^2 }. \tag*{\hfill\qedsymbol} \nonumber
\end{equation}

Note that the peak of the derivative of $G(\phi)$ and the sigmoid function $\sigma(\cdot)$ may not coincide, and 
as a result we expect the Lipschitz constant achieved in practice to be smaller.

\section{Model Architecture} \label{sec:appendix:model}
All the learned models described in this paper (Ours, \textit{AngleInput}, and \textit{AngleFree}) predict categorical distributions composed of $B=8$ bins, parameterized by the concentration parameters $[\gamma_\phi^1, \dots, \gamma_\phi^8]$.

All models share the same Point Pillars encoder
architecture~\citep{lang2019pointpillars}, which extracts a feature
representation from the input point cloud $\mathbf{Q}$. 
Specifically, the point cloud $\mathbf{Q}$ is first uniformly scaled to fit within 
a cube defined by $x \in [-0.5, 0.5]$, $y \in [-0.5, 0.5]$, and $z \in [0, 0.5]$. 
This normalized point cloud is then discretized into uniformly spaced pillars in the x-y plane,
each with a grid size of $0.1 m$. 
Points within each pillar are augmented with $x_c, y_c, z_c, x_p$, and $y_p$ where the 
$c$ subscript denotes distance to the arithmetic mean of all points in the pillar and the
$p$ subscript denotes the offset from the pillar's $x,y$ center~\citep{lang2019pointpillars}.
Notice we exclude the reflectance feature, yielding an 8-dimensional augmented input
representation for each lidar point. 
For each pillar, the encoder samples up to 32 points and generates a corresponding 32-dimensional feature vector. 
The resulting output from the Point Pillars encoder is a pseudo-image tensor with
dimensions $10 \times 10 \times 32$.

Next, we detail the remaining components of the model architecture.

\subsection{Neural Network Used in SPARTA}
The pseudo-image from the Point Pillars encoder~\citep{lang2019pointpillars} is subsequently processed by a two-layer convolutional backbone, specified as follows:
\begin{itemize}
    \item Input Channels: [32, 64]
    \item Output Channels: [64, 256]
    \item Kernel Size: [3, 3]
    \item Stride: [2, 2]
    \item Padding: [1, 1]
\end{itemize}

Each convolutional layer in the backbone employs batch normalization and ReLU activation. The resulting output tensor, of dimensions $3 \times 3 \times 256$, is passed through a max pooling operation to yield a compact 256-dimensional point cloud feature representation.

To represent the smooth analytical mapping $\phi \rightarrow \gamma_\phi^i$, we employ Fourier basis functions up to a maximum frequency of $n=3$, specifically the basis set $\{1, \cos(\phi), \sin(\phi), \cos(2\phi), \sin(2\phi), \cos(3\phi), \sin(3\phi)\}$. 
The choice of $n$ is justified in \cref{sec:appendix:open_loop} with an ablation study.
Consequently, for each bin $i \in [1, 8]$, the network predicts 7 Fourier coefficients. Thus, the SPARTA decoder consists of a multilayer perceptron (MLP) that transforms the 256-dimensional point cloud features into a 56-dimensional output vector $[\mathbf{F}^1, \dots, \mathbf{F}^8]$. This MLP decoder contains one hidden layer with 512 neurons and utilizes layer normalization and ReLU activation functions.

\subsection{AngleInput}
The pseudo-image from the Point Pillars encoder~\citep{lang2019pointpillars} is subsequently processed by a three-layer convolutional backbone, specified as follows:
\begin{itemize}
    \item Input Channels: [32, 64, 128]
    \item Output Channels: [64, 128, 256]
    \item Kernel Size: [3, 3, 3]
    \item Stride: [2, 2, 2]
    \item Padding: [1, 1, 0]
\end{itemize}
The encoder for angle of approach $\phi$ is a MLP that maps 1-dimensional input $\phi$ to a 32-dimensional embedding with one hidden layer with 16 neurons.
The decoder merges the point cloud feature (256-dimensional) and the angle feature (32-dimensional), and outputs the 8 concentration parameters $[\gamma_\phi^1, \dots, \gamma_\phi^8]$, with a hidden layer with 64 neurons.
Both MLPs utilizes layer normalization and ReLU activation functions.

\subsection{AngleFree}
The \textit{AngleFree} baseline network uses the same convolutional backbone as \textit{AngleInput}.
A decoder head transforms the 256-dimensional point cloud feature to the 8 concentration parameters $[\gamma_\phi^1, \dots, \gamma_\phi^8]$, with two hidden layers with [128, 64] neurons.

\section{Inference Efficiency} \label{sec:appendix:runtime}
\begin{wrapfigure}[18]{r}{0.35\textwidth}
    \centering
    \vspace{-1.2em}
    \includegraphics[width=\linewidth]{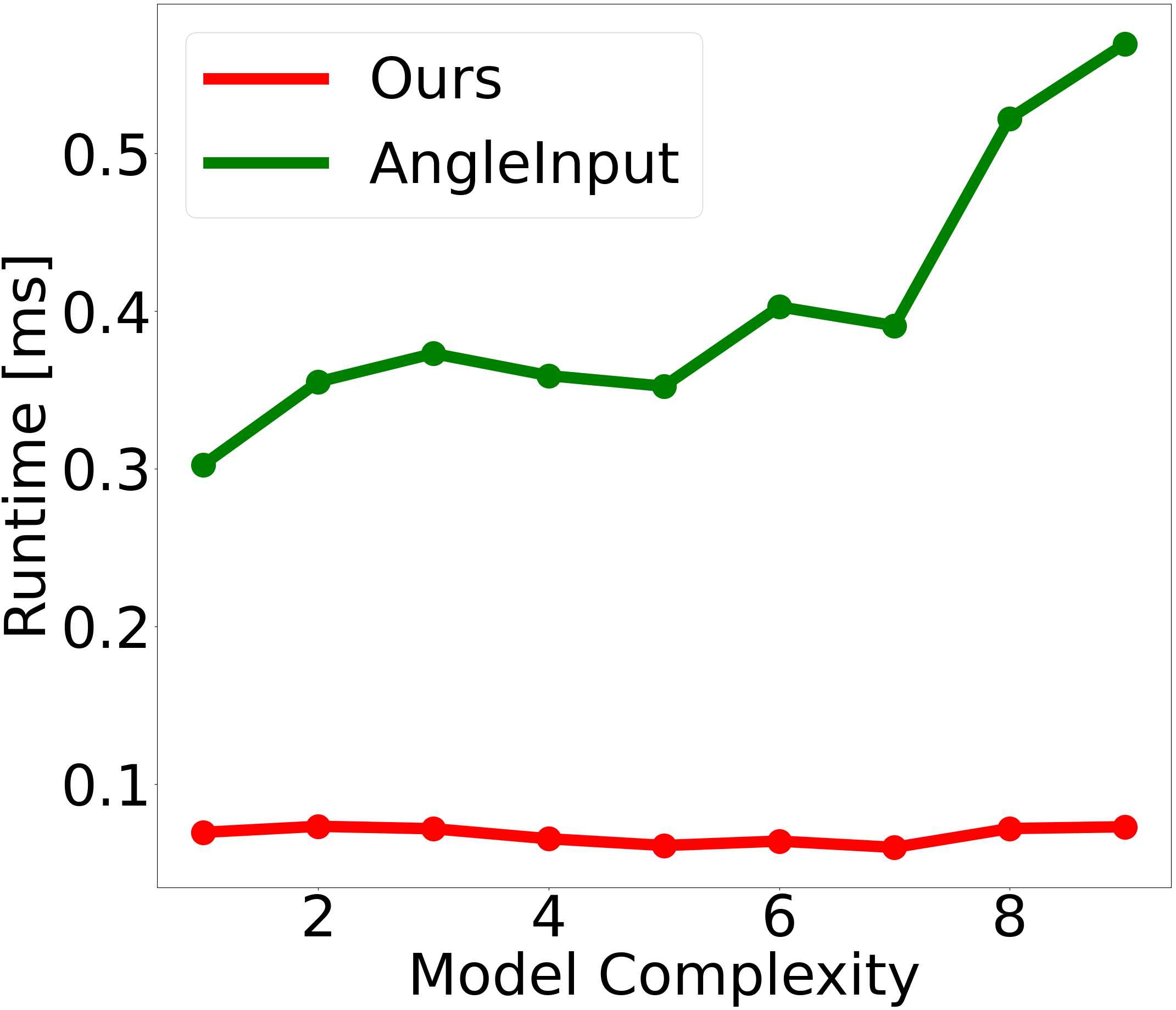}
    \vspace{-10pt}
    \caption{
        Runtime of \textit{AngleInput} (Green) increases with model complexity,
        while the runtime of ours (Red) remains roughly constant. This is because during planning, we only need 
        to perform a low dimensional dot product and a sigmoid pass during inference, whereas \textit{AngleInput} 
        requires repeated neural network inference. The x-axis denotes additional hidden layers in the decoder 
        and additional frequencies in our model, i.e., $n \in [3, 13]$.
    }
    \label{fig:appendix:runtime}
\end{wrapfigure}

% To demonstrate the query efficiency of our model compared to \textit{AngleInput}, we measure the inference
% runtime of the two models.
% %
% Specifically, for \textit{AngleInput} we measure the runtime for encoding the approach angle and pass through the decoder head for
% the concentration parameters, and for our model we measure the runtime for computing \cref{eqn:fourier}, as other parts
% of the models can be precomputed/re-used.
% %
% When dealing with more complex point clouds, the point cloud feature dimension and the number of model layers
% will have to increase for performance.
% %
% Therefore we stress test the inference time by gradually increasing the number of layers and width of the decoder.
% %
% We additionally increase the maximum frequency $n$ in our model.
% %
% We run both models to process 25000 queries (an estimated number of queries needed per planning step)
% and report the average runtimes in \cref{fig:appendix:runtime}
% \footnote{The x-axis denotes additional hidden layer in decoder, and also denotes additional Fourier 
% basis functions used in our model, i.e. using [3, 13] Fourier basis functions.}.
% %
% The runtime of our model is an order-of-magnitude faster than that of \textit{AngleInput}.
% %
% Most importantly, our model's runtime remains constant with increasing model complexity, because we only need a single batched,
% low dimensional dot product and sigmoid, whereas the runtime of \textit{AngleInput} significantly increases.\par

To demonstrate the query efficiency of our model compared to \textit{AngleInput}, we measure their respective inference runtimes. 
Specifically, for \textit{AngleInput}, we measure the runtime of encoding the approach angle, decoding the concentration parameters,
and computing the categorical distribution (\cref{eqn:PMF}).
For our model, we measure the runtime associated only with computing the concentration parameters (\cref{eqn:fourier})
and the categorical distribution (\cref{eqn:PMF}).
Note other components (e.g. extracting feature vector from point cloud) of both models can be precomputed or reused.
Considering that more complex point clouds require higher-dimensional point cloud features and deeper network architectures,
we perform a stress test by progressively increasing the number of layers in both models,  as well as the maximum frequency $n$ in our model. 
Both models process 25,000 queries with average runtimes reported in \cref{fig:appendix:runtime}.
% \footnote{The x-axis denotes additional hidden layers in the decoder and additional Fourier basis functions in our model, i.e., ranging from [3, 13] basis functions.}.
%
Our model demonstrates an order-of-magnitude faster runtime compared to \textit{AngleInput}. 
Most importantly, the runtime of our model remains constant despite increased complexity, 
as it only involves a single batched, low-dimensional dot product and sigmoid computation
\footnote{The dimension of the dot product depends only on the choice of maximum frequency $n$. In fact
the dimension is exactly $2n+1$.}. 
In contrast, the runtime of \textit{AngleInput} increases with complexity.\par

\section{Data Collection Detail} \label{sec:appendix:data}
In this section we provide more details of our data collection process using the
automotive simulator \textit{BeamNG.tech}~\citep{beamng_tech}.
We intend to generate random obstacles for the vehicle to drive over, and collect the 
tire deformation during the traversal.
We choose a patch size of $1 \times 1m$ that covers the footprint of the vehicle wheel,
and randomly place (non-overlapping) obstacles in the patch.
Maximum elevation of the obstacle is set to $0.5m$ (the vehicle's minimum ground clearance).
For every patch, we collect its point cloud and downsample to a maximum of 1024 points.
To minimize the influence of vehicle dynamics (for the purpose of generalization to unseen hardware), we
% let
place the vehicle such that only one wheel contacts the obstacle.
The deformation extent $d_w = \frac{r_w - r_\text{inner}}{r_\text{outer} - r_\text{inner}}$
of all contacting tire node $w$ are recorded during the traversal.
We discard all samples below 0.2 (the tire deformation when the vehicle is static on flat ground) 
and divide the rest into a histogram with 8 bins, which forms the empirical distribution $\mathbf{y}$.
In total, 24,000 samples $\{\mathbf{Q}_k, \mathbf{y}_k, \phi_k\}$ are collected.
We augment the samples by applying 8 random rotations (to both the point cloud and the angle of approach),
resulting in a total of 216,000 samples.
We further add random perturbation to the point clouds for more robust performance.
We use 90\% of the data for training and 10\% for testing.

\section{Traversing a Row of Obstacles with Ablation Study} \label{sec:appendix:open_loop}
% \begin{wraptable}{r}{0.4\textwidth}
%     \centering
%     \vspace{-0.15in}
%     \footnotesize
%     \begin{tabular}{c | c | c c}
%         \hline
%         Algorithm & $\alpha$ & S.R. $\uparrow$ & F (N) $\downarrow$ \\
%         \hline
%         \multirow{2}{2em}{Ours} &   0.9 &   \textbf{95}    &   \underline{450.19}  \\
%                                 &   0   &   89             &   \textbf{432.26}     \\
%         \hline
%         \multirow{2}{4em}{\textit{AngleInput}}  &   0.9 &   \underline{92} &   492.61              \\
%                                 &   0   &   89             &   548.91              \\
%         \hline
%         \textit{Elev}                    &   --  &   73             &   740.04              \\
%         \hline
%     \end{tabular}
%     \caption{Success Rate (S.R.) and Average Force (F) experienced by wheel nodes in simulation. Our model outperforms 
%     the baselines \textit{AngleInput} and \textit{Elev}. The performance of the models decrease when we ablate CVaR with mean. Best
%     result is \textbf{Bolded}, and second best is \underline{underlined}.}
%     \label{tab:exp:obs_belt}
%     \vspace{-0.15in}
% \end{wraptable}

% \textbf{TODO: Example of the environment (Screenshot), example point clouds, with predicted CVaRs}
\begin{figure}[t]
    \centering
    \vspace{-1.2em}
    \begin{subfigure}[t]{0.24\linewidth}
        \centering
        \includegraphics[width=\linewidth]{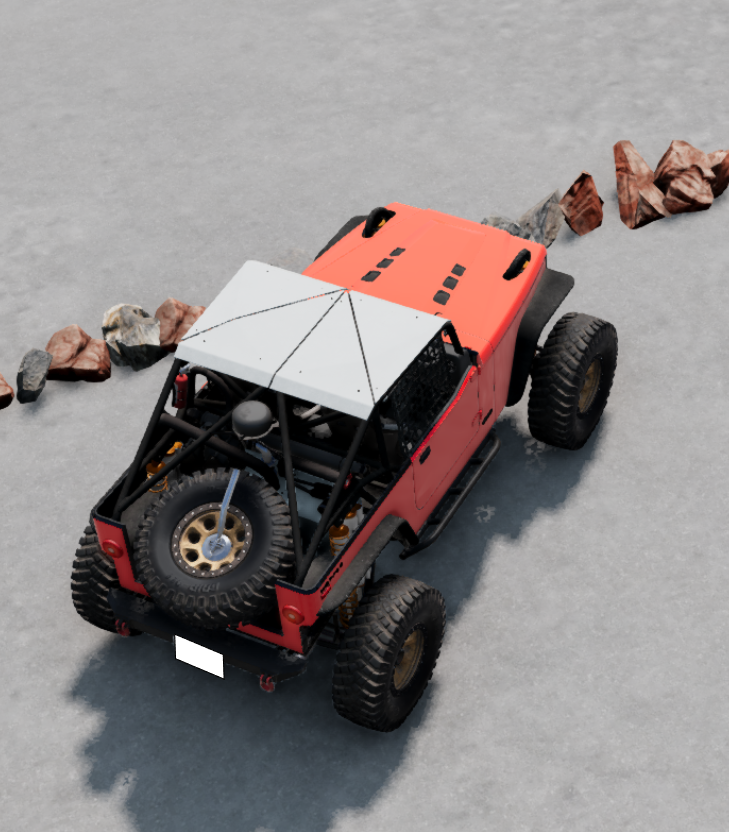}
        \captionsetup{skip=0pt}
        \caption{}
        \label{fig:appendix:row:setting}
    \end{subfigure}
    \begin{subfigure}[t]{0.24\linewidth}
        \centering
        \includegraphics[width=\linewidth]{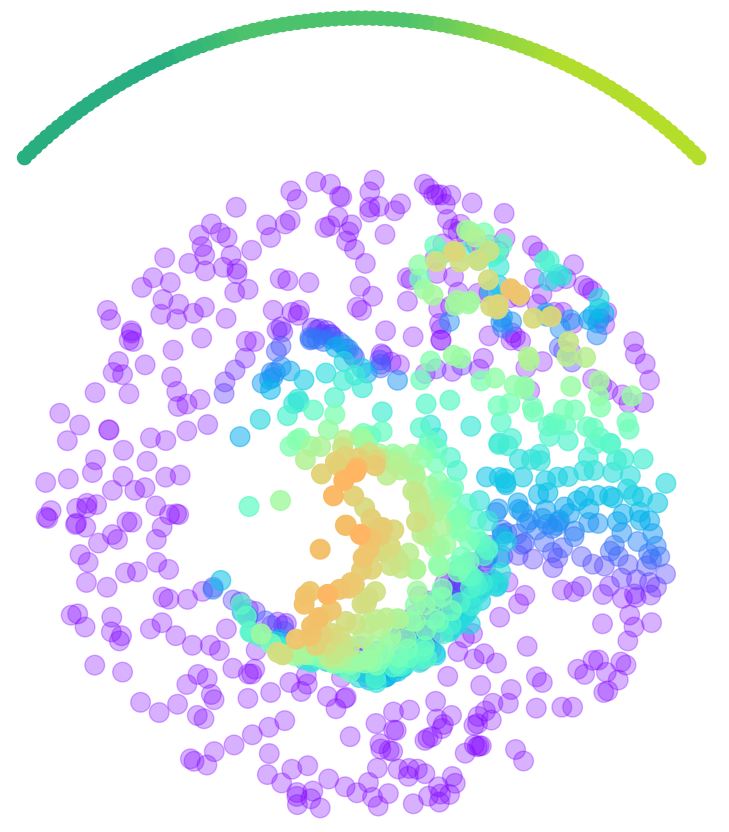}
        \captionsetup{skip=0pt}
        \caption{}
        \label{fig:appendix:row:eg1}
    \end{subfigure}
    \begin{subfigure}[t]{0.24\linewidth}
        \centering
        \includegraphics[width=\linewidth]{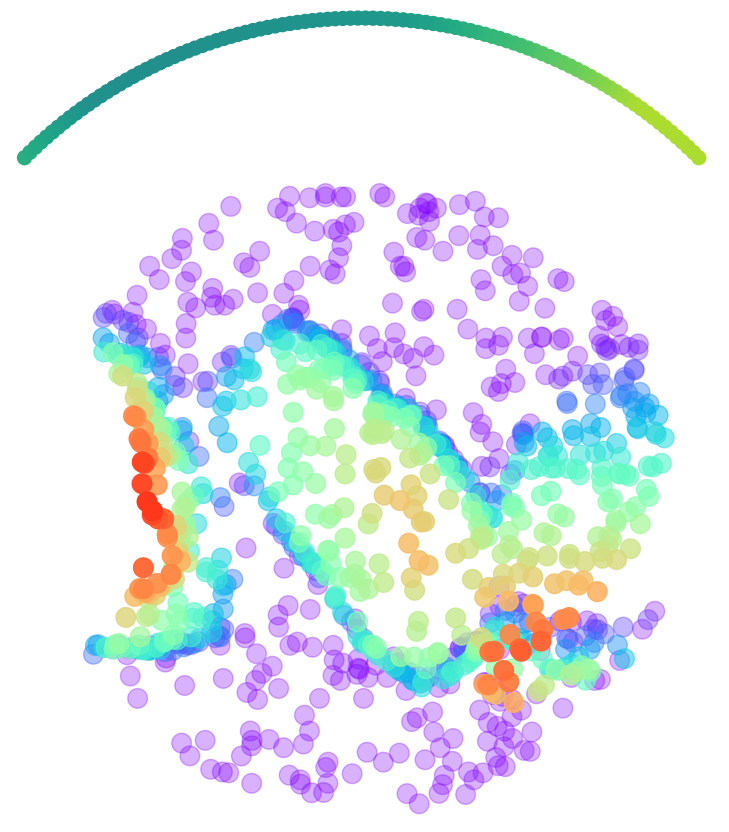}
        \captionsetup{skip=0pt}
        \caption{}
        \label{fig:appendix:row:eg2}
    \end{subfigure}
    \begin{subfigure}[t]{0.24\linewidth}
        \centering
        \includegraphics[width=\linewidth]{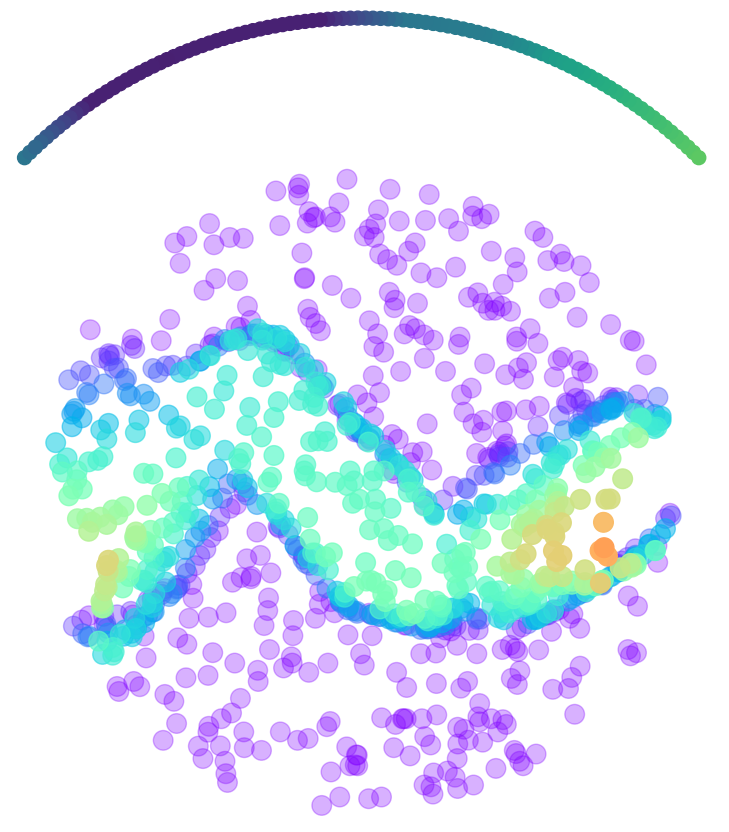}
        \captionsetup{skip=0pt}
        \caption{}
        \label{fig:appendix:row:eg3}
    \end{subfigure}
    \begin{subfigure}[t]{0.35\linewidth}
        \centering
        \includegraphics[width=\linewidth]{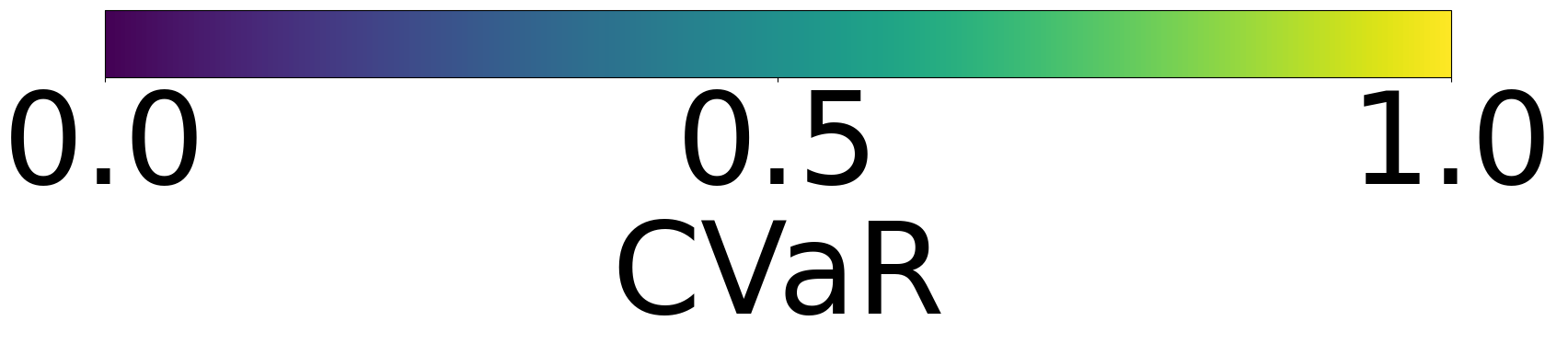}
        \captionsetup{skip=0pt}
    \end{subfigure}
    \hspace{0.2\linewidth}
    \begin{subfigure}[t]{0.35\linewidth}
        \centering
        \includegraphics[width=\linewidth]{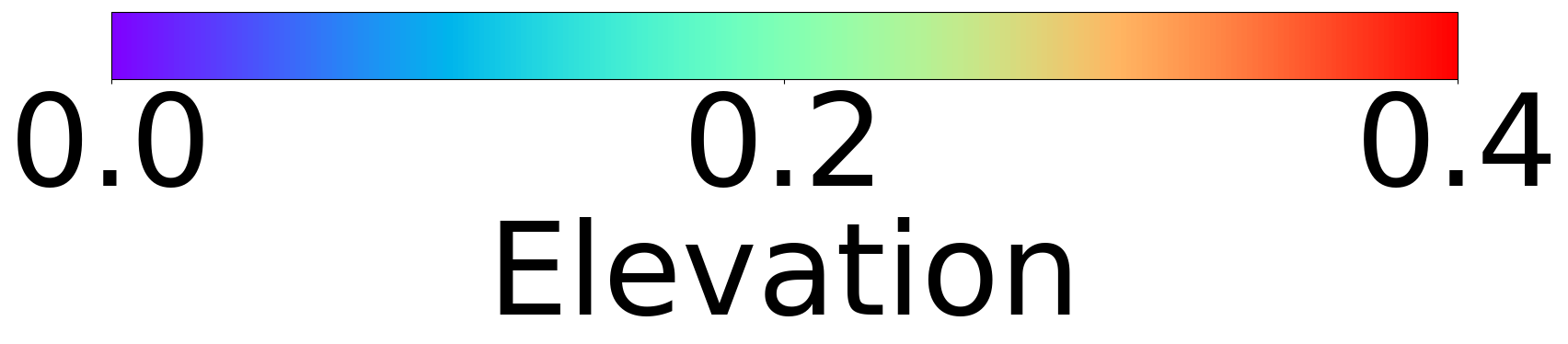}
        \captionsetup{skip=0pt}
    \end{subfigure}
    \caption{
        (a): Overview of the test environment and the vehicle.
        (b)-(d): Example point cloud (colored by elevation) and the CVaR (Viridis arces around the point clouds)
        predicted by our model at the candidate angles of approach. 
    }
    \label{fig:appendix:row}
\end{figure}

\begin{wraptable}{r}{0.4\textwidth}
    \centering
    \vspace{-1.2em}
    \footnotesize
    \begin{tabular}{c | c | c | c c}
        \hline
        Algorithm & $\alpha$ & $n$ & Suc. $\uparrow$ & Dmg. $\downarrow$ \\
        \hline
        \multirow{4}{2em}{Ours} &   0.9 &  3  &  \textbf{95}    &   \textbf{5}              \\
                                &   0.9 &  1  &  91             &   9                       \\
                                &   0.9 &  5  &  \underline{94} &   \underline{6}           \\
                                &   0   &  3  &  89             &   11                      \\
        \hline
        \multirow{2}{4em}{AngleInput}  &   0.9  &  --  &  92  &  8             \\
                                       &   0    &  --  &  89  &  11            \\
        \hline
        Elev                           &   --   &  --  &  73  &  27            \\
        \hline
    \end{tabular}
    \caption{Number of success (Suc.) and vehicle damage (dmg.) in simulation. Our model outperforms 
    \textit{AngleInput} and \textit{Elev}. The performance of the models decrease when we ablate CVaR with mean. \textbf{Best}
    , \underline{Second best}.}
    \label{tab:exp:obs_belt}
    \vspace{-0.15in}
\end{wraptable}
In this experiment, we test our model's ability to identify safe traversal points and angles of approach
in a row of packed, randomly placed obstacles.
An overview of the environment, a few example point clouds (colored by elevation), and the CVaRs predicted by
our model are shown in \cref{fig:appendix:row}.
This environment is simpler than the boulder field in \cref{sec:exp:sim} and thus helpful in isolating
the influence of the variables we are interested in.
We consider two baselines, the \textit{AngleInput} and a heuristic baseline that chooses the point with lowest maximum elevation
under vehicle's wheel footprint (\textit{Elev}).
Since the obstacles are placed on flat ground, the \textit{Elev} baseline is equivalent to finding the plan with
lowest stepping difficulty as in~\citep{fan2021step, lee2025trg}.
% %
% We spawn a row of packed, randomly rotated and scaled obstacles in a narrow passage way between
% two walls (lethal obstacle), and thus force the vehicle to traverse the obstacle row in order to
% reach the goal placed on the other side of the obstacles.
% %
% The size of the obstacles is set to approximately $0.5m \times 0.5m \times 0.35m$ (slightly larger than 
% vehicle tire width and lower than vehicle ground clearance).
% %
% The grid resolution for elevation map and precomputed Fourier coefficients are set to 0.2m
% for all experiments in simulation.
% %
% A successfully traversal means the vehicle crossed the obstacles without damaging, and reaches
% the goal within the maximum steps allowed.
% %
% Besides the \textit{AngleInput} baseline, we include another baseline that selects the point/angle that induces
% the lowest elevation below the vehicle's wheel footprint (\textit{Elev}).
%
As shown in \cref{tab:exp:obs_belt}, our model achieves higher success rate (S.R., the vehicle drives over the obstacle without
damage) than the \textit{AngleInput}, but both learning-based method significantly outperforms \textit{Elev}.
This is expected because the obstacles are spawned with similar sizes, and as a result \textit{Elev} degenerates to random selection, 
leading to much lower success rate and thus high risk of vehicle damage.
% %
% As mentioned in \cref{sec:exp:data_collection}, we choose to predict tire deformation as a surrogate to vehicle damage.
% %
% The validity of this choice is demonstrated in the results because high risk of vehicle damage correlates with
% higher force, i.e., we observe the \textit{Elev} baseline induces much higher force on wheel nodes compared to the 
% other candidates.

\paragraph{Ablating CVaR} 
To demonstrate the effectiveness of using CVaR during planning to capture worst-case risk, we ablate it
by setting $\alpha=0$, which is equivalent to using the mean of the predicted distribution.
As reported in \cref{tab:exp:obs_belt}, the performance of both our model and \textit{AngleInput} dropped.
This is because accounting for worst-case (tail) risk is crucial in safety-critical tasks, and CVaR is a principled way for
representing tail risk in robotics~\citep{majumdar2020should}.
% This is because the CVaR accounts for tail risk distribution and is thus more suitable for safety-critical tasks.

\paragraph{Ablating Maximum Frequency}
\begin{wrapfigure}[11]{r}{0.3\textwidth}
    \centering
    % \vspace{-1.2em}
    \includegraphics[width=\linewidth]{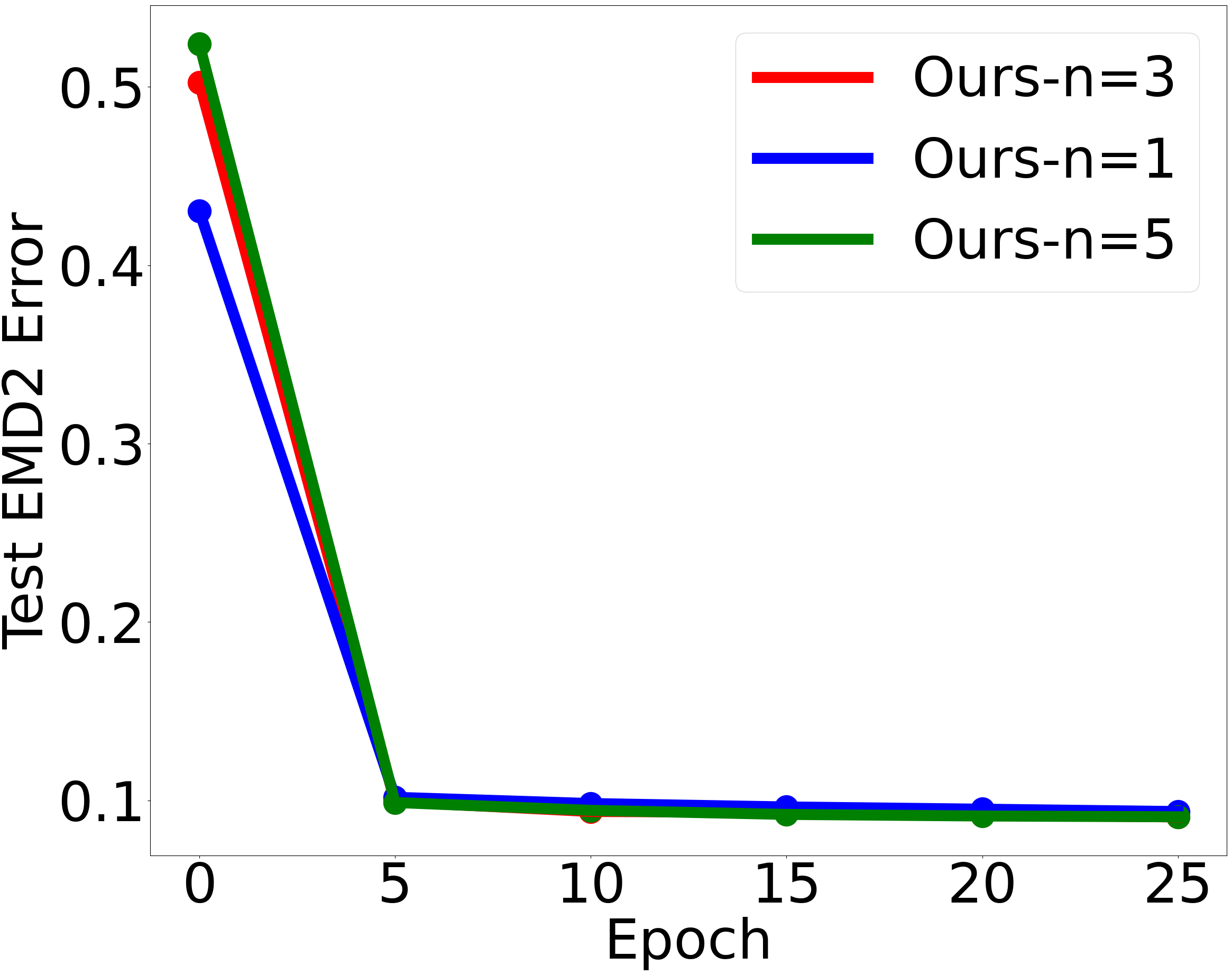}
    \caption{
        Test $\text{EMD}^2$ loss of our model with varying maximum frequency. They achieve similar test loss.
    }
    \label{fig:appendix:ablate_n}
\end{wrapfigure}
As demonstrated in \cref{sec:method:representation} and \cref{sec:method:theory}, the choice of maximum
frequency $n$ is closely related to the smoothness and generalization of the mapping from angle of approach $\phi$
to each of the concentration parameters $\gamma_\phi^i$.
Therefore, we adjust the number of Fourier bases $n$ in \cref{eqn:fourier} and analyze the results in this section.
We consider the cases $n=1$ and $n=5$, and visualize the test $\text{EMD}^2$ loss in \cref{fig:appendix:ablate_n}.
Overall, we observe similar test $\text{EMD}^2$ loss for all three models.
%, demonstrating that our model
%is capable of generalizing robustly in the test set despite the choice of maximum frequency $n$.
%
In \cref{tab:exp:obs_belt} we include their performances on the aforementioned experiment, on which
they achieve success rate 91\% and 94\%, respectively.
Notice the model with $n=1$ achieves lower success rate compared to the other two models.
We hypothesize this is because the basis functions can be too simple to capture the change of $\gamma_\phi^i$ with $\phi$, i.e.
it might not be expressive enough to parameterize the mapping we are interested in.
Although the model with maximum frequency $n=5$ achieves similar success rate as ours ($n=3$), an overly high maximum frequency
$n$ introduces higher upper bound on the Lipschitz constant of the mapping and can introduce more abrupt changes in $\gamma_\phi^i$
with small changes in $\phi$, as explained in \cref{sec:method:theory}.
Therefore, to balance expressivity and generalization, we choose $n=3$ in our
model and our experiments. However, $n$ is an easily changed hyperparameter that
can be adjusted depending on the frequency composition of the mapping function
we are trying to approximate.
\par

\section{Hardware Specification} \label{sec:appendix:hardware}
Our robot is built on top of the AgileX Scout Mini chassis, a skid-steering wheeled platform.
It has a footprint of approximately $0.6 \times 0.6m$ and can achieve a maximum speed of 3 m/s.
The robot is equipped with an Ouster OS1 LiDAR with 128 channels.
The LiDAR runs with a horizontal resolution of 1024 at 10 Hz, and we use its internal 6-axis
IMU (100 Hz).
The point cloud and IMU measurements are processed using DLIO~\citep{chen2023dlio} for prebuilding
the point cloud map and online localization.
The onboard PC is an ASUS ROG NUC 970, with an Intel Core Ultra 9 CPU (16-Core), 32 GB RAM,
and a Nvidia GeForce RTX 4070 GPU (8 GB).

% \begin{itemize}
%     \item Chassis: Agile X Scout Mini
%     \item Onboard Sensors:
%     \begin{itemize}
%         \item LiDAR: Ouster OS-1 128 Channel, operating with 1024 horizontal resolution at 10 Hz.
%         \item IMU: Ouster OS-1 Internal IMU (6-axis), operating at 100 Hz.
%     \end{itemize}
%     \item Onboard PC: ASUS ROG NUC 970
%     \begin{itemize}
%         \item CPU: 2.3 GHz Intel Core Ultra 9 185H 16-Core
%         \item RAM: 32 GB 5600 MHz DDR5 
%         \item GPU: Nvidia GeForce RTX 4070 (8GB GDDR6)
%     \end{itemize}
% \end{itemize}

\section{Generalization to Other Problem Settings} \label{sec:appendix:example}
\begin{wrapfigure}[20]{r}{0.4\textwidth}
    \centering
    \vspace{-1.2em}
    \begin{subfigure}[t]{\linewidth}
        \includegraphics[width=\linewidth]{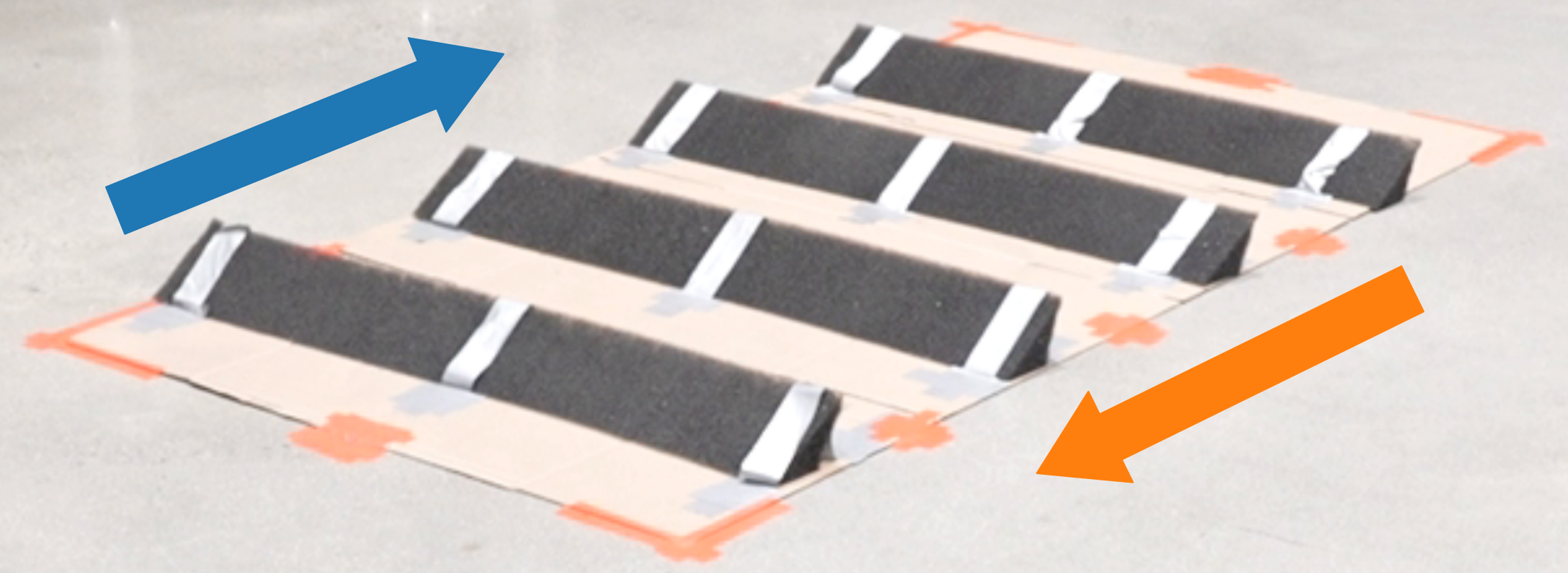}
        \caption{} \label{fig:appendix:traction:a}
    \end{subfigure}
    \begin{subfigure}[t]{\linewidth}
        \includegraphics[width=\linewidth]{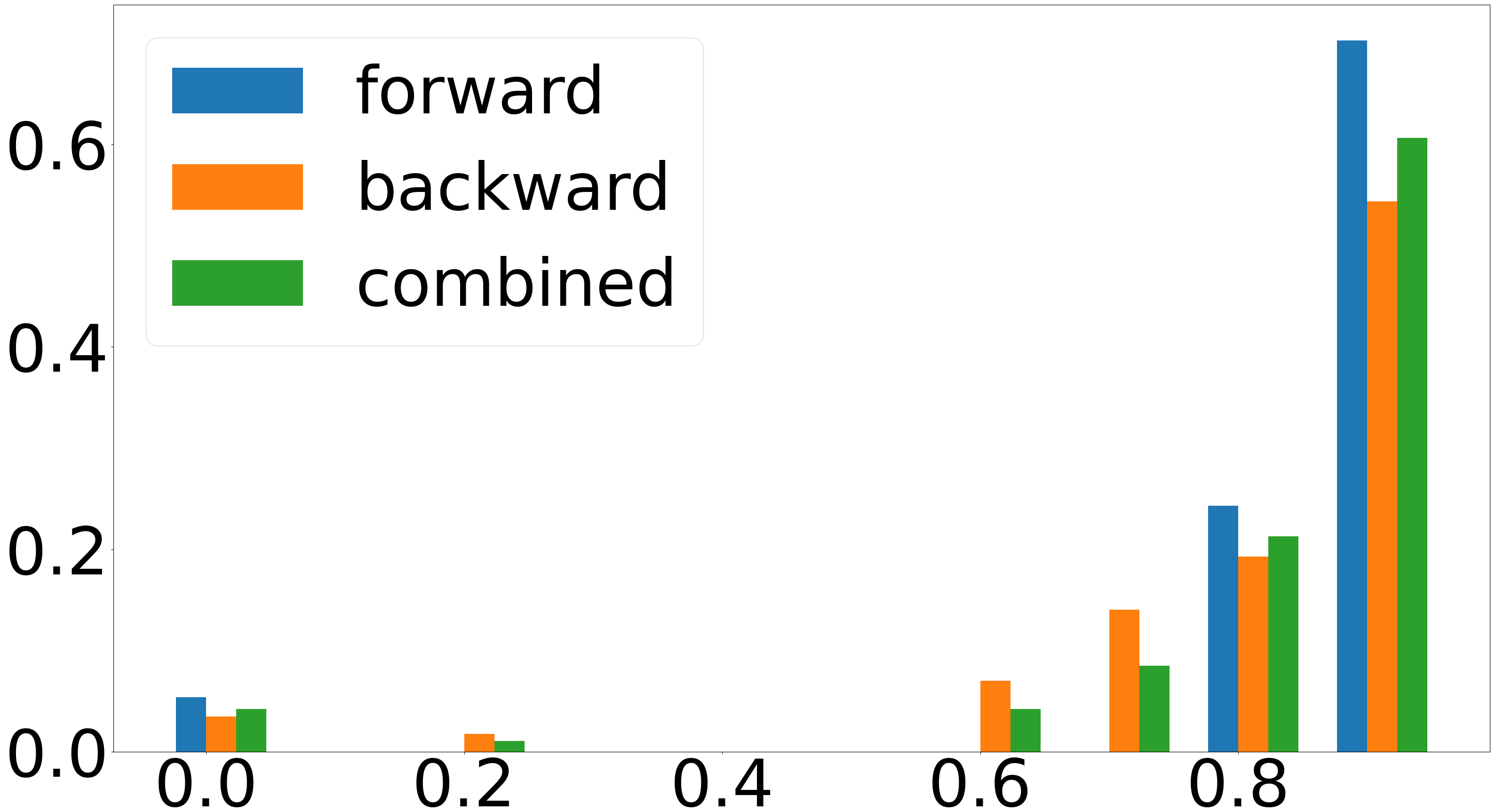}
        \caption{} \label{fig:appendix:traction:b}
    \end{subfigure}
    \caption{
        (a): Obstacle and directions of the traversals.
        (b): The linear traction distribution of the forward (Blue) and backward (Orange) traversal. The
        backward distribution has significantly more probability mass in the lower bins. An algorithm
        not modeling the angle-dependency of linear traction (Green) would not be able to identify the obstacle
        as traversable from the forward orientation.
    }
    \label{fig:appendix:traction}
\end{wrapfigure}
In this section, we first demonstrate other problem settings that could benefit from modeling angle-dependency.
We use the angle-dependency of linear traction (the robot's ability to
execute commanded velocity on the given terrain~\citep{cai2024evora}) as an example,
and show how our approach can be applied to this problem setting.
To construct an empirical dataset, we command the vehicle to drive at a constant
linear velocity of 0.5 m/s over the obstacle, along the forward (Blue) and
the backward (Orange) direction shown in \cref{fig:appendix:traction:a}.
We record the achieved linear velocity (estimated via DLIO~\citep{chen2023dlio}), and compute
the linear traction as the ratio between the achieved linear velocity and the commanded linear
velocity (0.5 m/s).
The data points are discretized into histograms and then normalized into categorical distributions.
As shown in \cref{fig:appendix:traction:b}, the backward traversal (Orange) achieved significantly lower
linear tractions compared to the forward traversal (Blue).
This matches our intuition because the obstacle is smooth from the forward direction and sharp from the other.
A system that models the angle-dependency of linear traction would thus be able to capture this difference.
Notice SPARTA can be applied to this problem setting with minimal changes (e.g. adjusting input point cloud size)
using an empirical dataset of linear traction distributions.
However, systems that do not model angle-dependency (Green)~\citep{cai2024evora} could be overly conservative
and never command the robot to traverse the obstacle from the forward direction, which is traversable in practice.

In fact, we anticipate any variables that exhibit smooth and continuous behaviors, especially those 
that depends on variables with Lie group structure, could similarly benefit from our representation framework. 
\textbf{SPARTA} can be easily modified to use their respective basis functions and predict the corresponding coefficients.

\section{Example CVaR Predictions of SPARTA} \label{sec:appendix:pred_example}
\begin{figure}[t]
    \label{fig:exp:crafted}
    \centering
    \vspace{-1.0em}
    \begin{subfigure}[t]{0.23\linewidth}
        \centering
        \includegraphics[width=\linewidth]{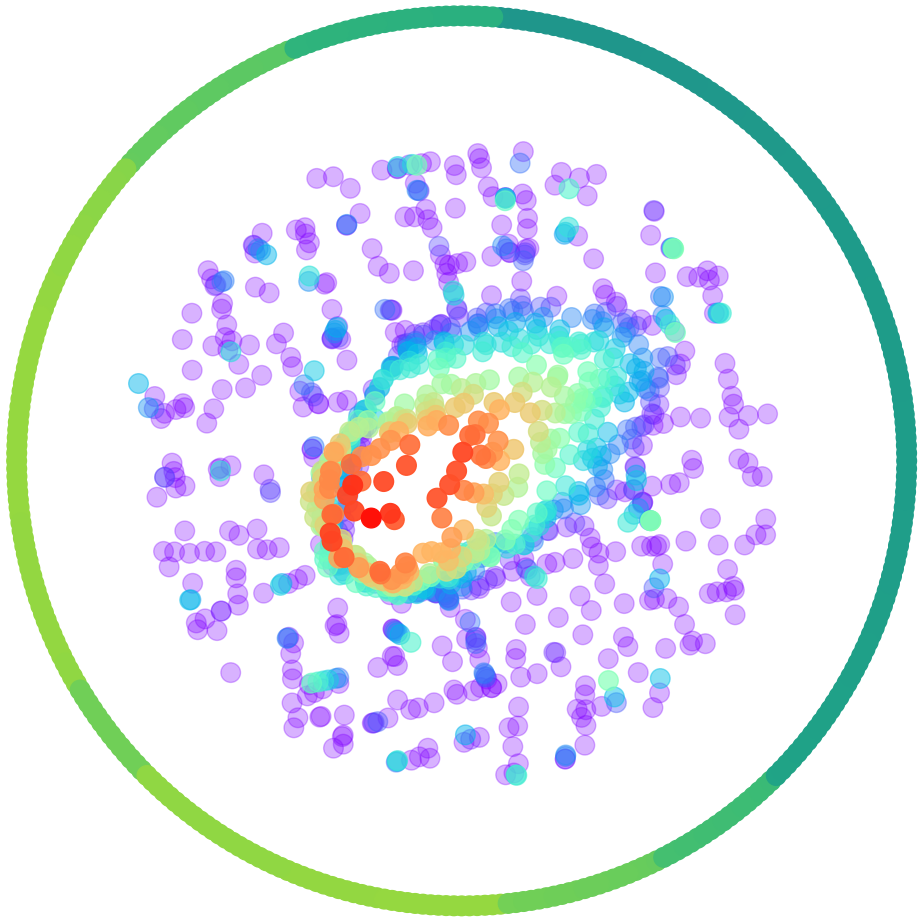}
        \captionsetup{skip=0pt}
        \caption{}
    \end{subfigure}
    \begin{subfigure}[t]{0.23\linewidth}
        \centering
        \includegraphics[width=\linewidth]{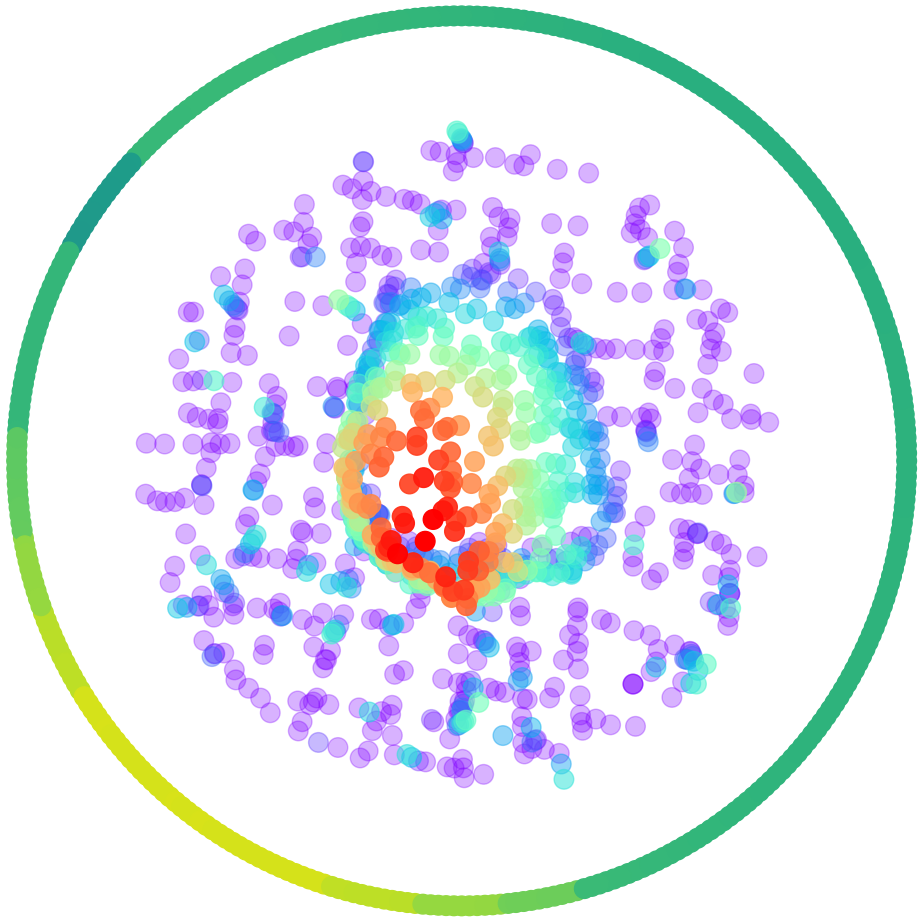}
        \captionsetup{skip=0pt}
        \caption{}
    \end{subfigure}
    \begin{subfigure}[t]{0.23\linewidth}
        \centering
        \includegraphics[width=\linewidth]{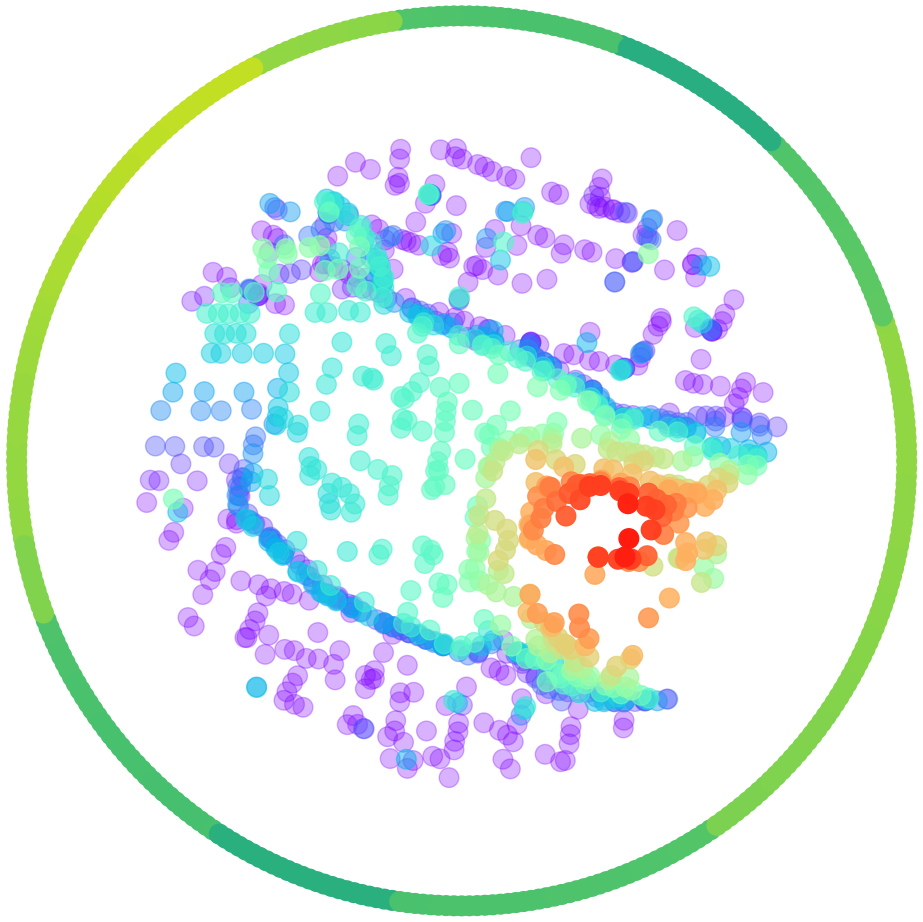}
        \captionsetup{skip=0pt}
        \caption{}
    \end{subfigure}
    \begin{subfigure}[t]{0.23\linewidth}
        \centering
        \includegraphics[width=\linewidth]{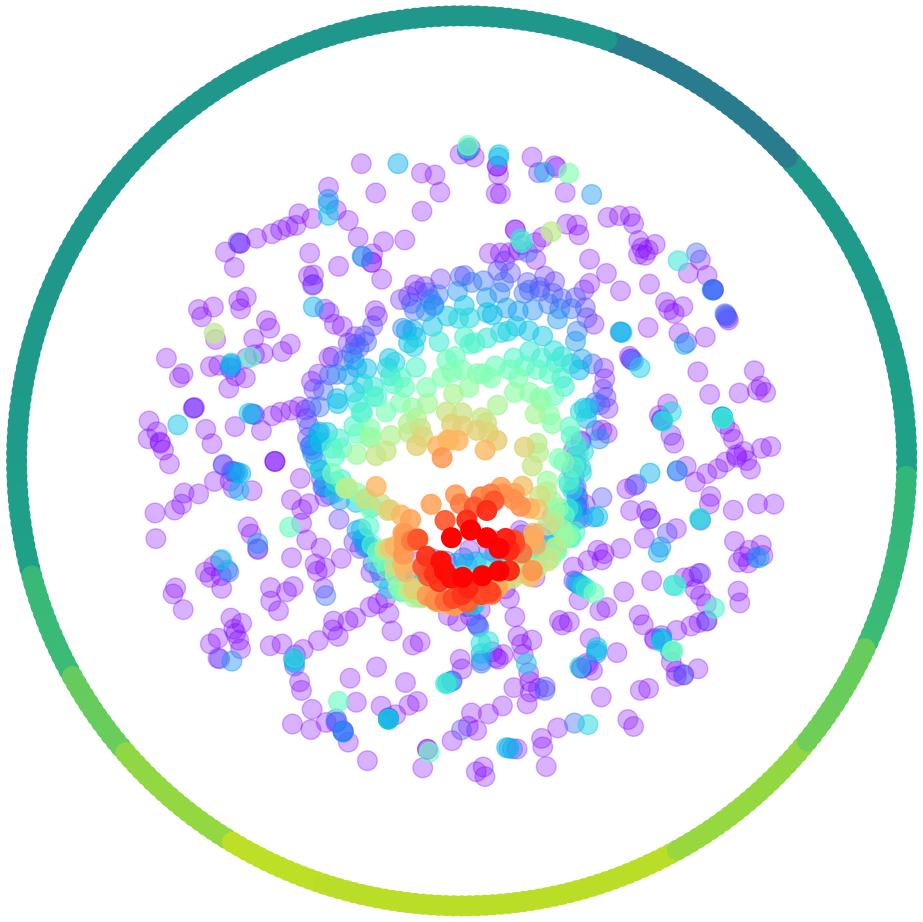}
        \captionsetup{skip=0pt}
        \caption{}
    \end{subfigure}
    \begin{subfigure}[t]{0.23\linewidth}
        \centering
        \includegraphics[width=\linewidth]{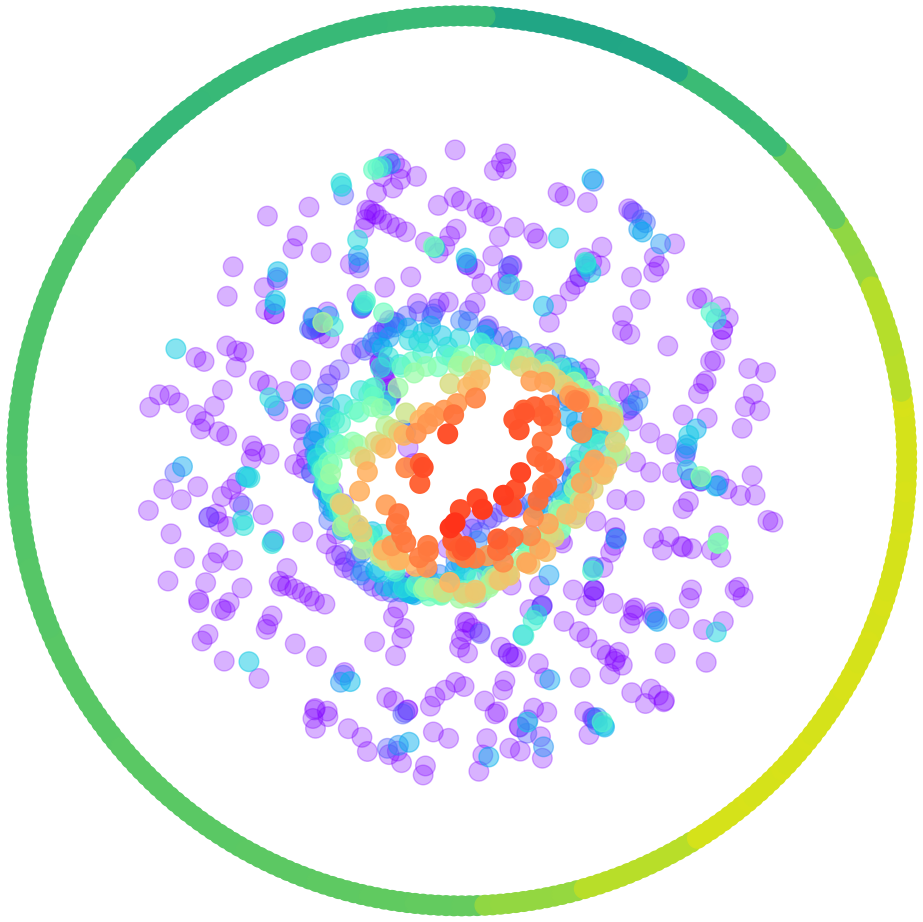}
        \captionsetup{skip=0pt}
        \caption{}
    \end{subfigure}
    \begin{subfigure}[t]{0.23\linewidth}
        \centering
        \includegraphics[width=\linewidth]{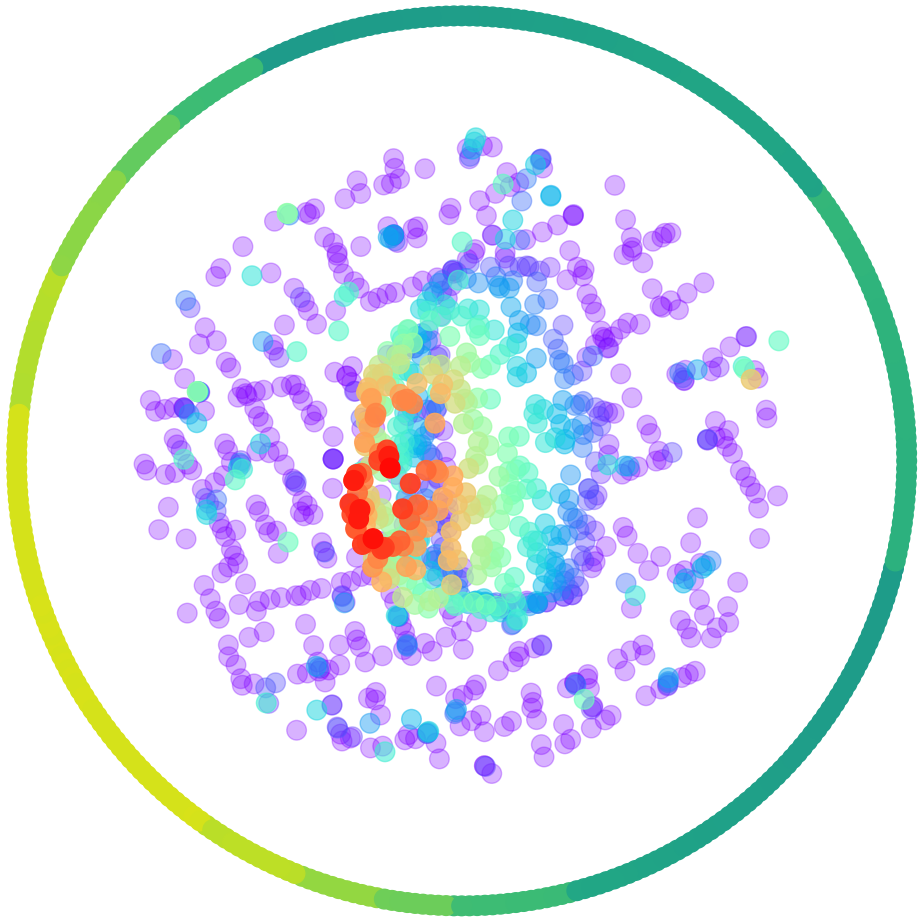}
        \captionsetup{skip=0pt}
        \caption{}
    \end{subfigure}
    \begin{subfigure}[t]{0.23\linewidth}
        \centering
        \includegraphics[width=\linewidth]{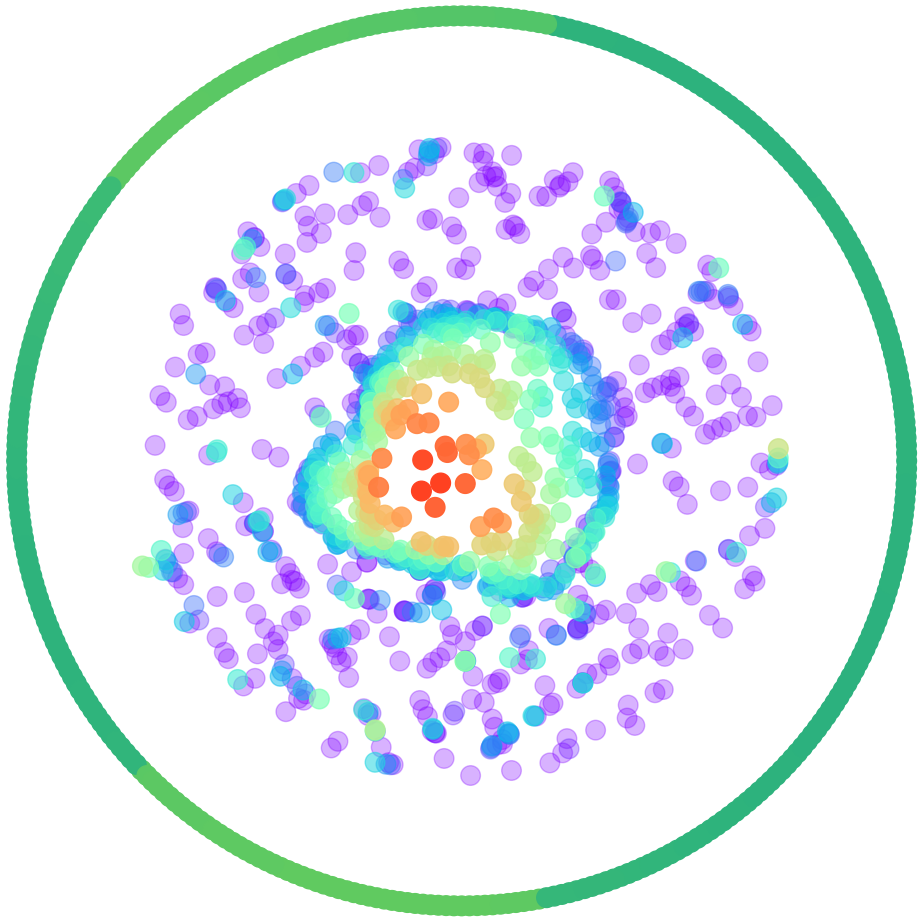}
        \captionsetup{skip=0pt}
        \caption{}
    \end{subfigure}
    \begin{subfigure}[t]{0.23\linewidth}
        \centering
        \includegraphics[width=\linewidth]{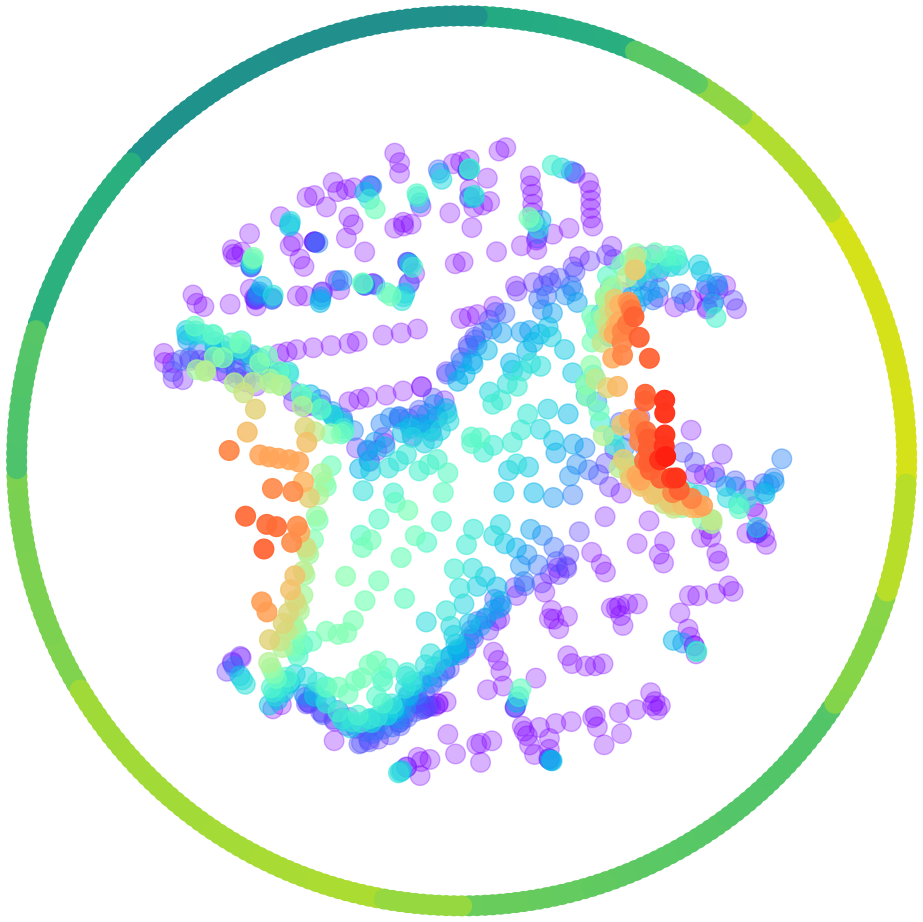}
        \captionsetup{skip=0pt}
        \caption{}
    \end{subfigure}
    \begin{subfigure}[t]{0.23\linewidth}
        \centering
        \includegraphics[width=\linewidth]{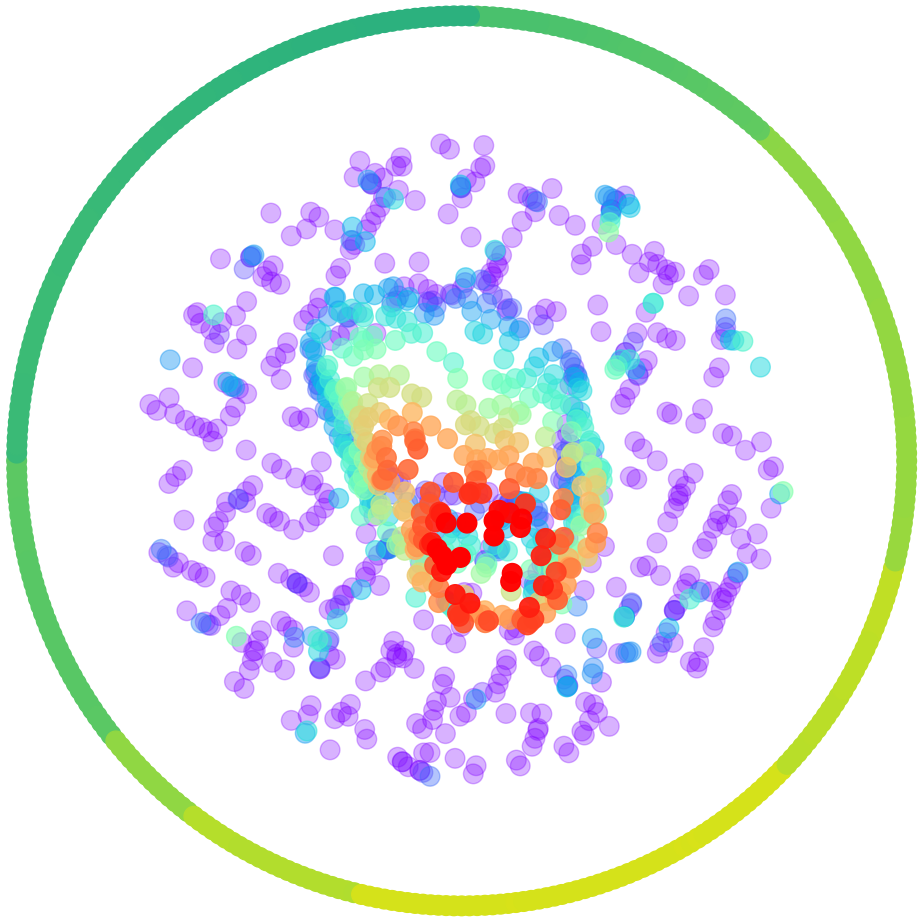}
        \captionsetup{skip=0pt}
        \caption{}
    \end{subfigure}
    \begin{subfigure}[t]{0.23\linewidth}
        \centering
        \includegraphics[width=\linewidth]{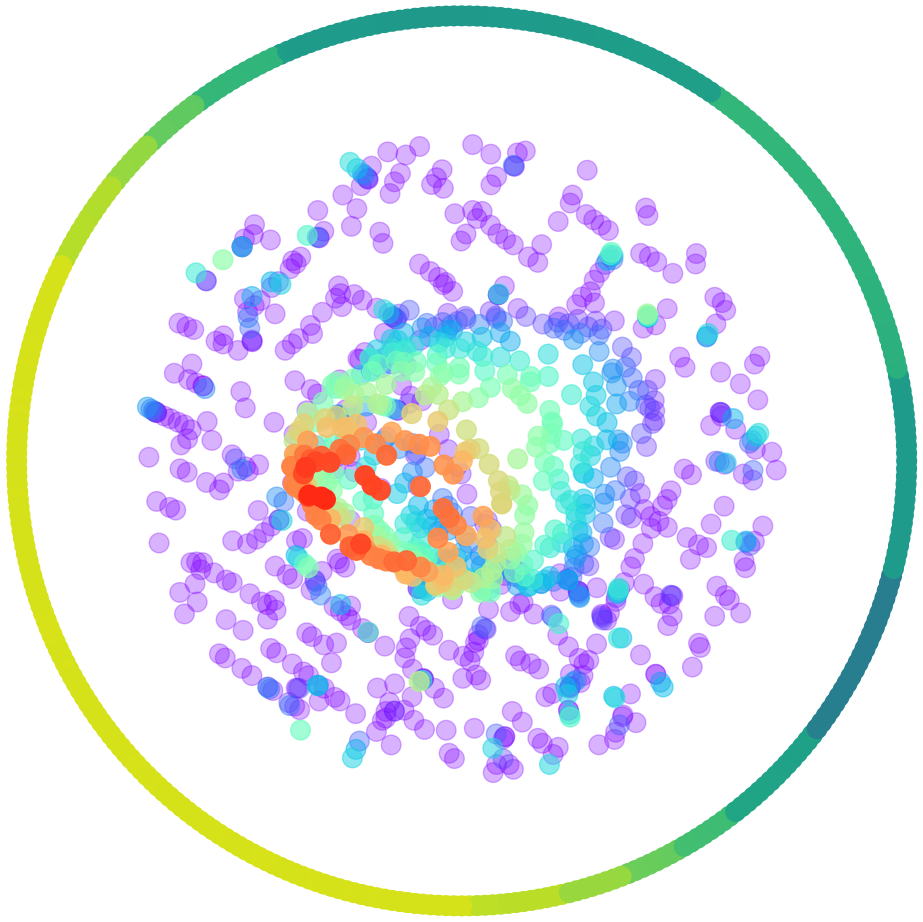}
        \captionsetup{skip=0pt}
        \caption{}
    \end{subfigure}
    \begin{subfigure}[t]{0.23\linewidth}
        \centering
        \includegraphics[width=\linewidth]{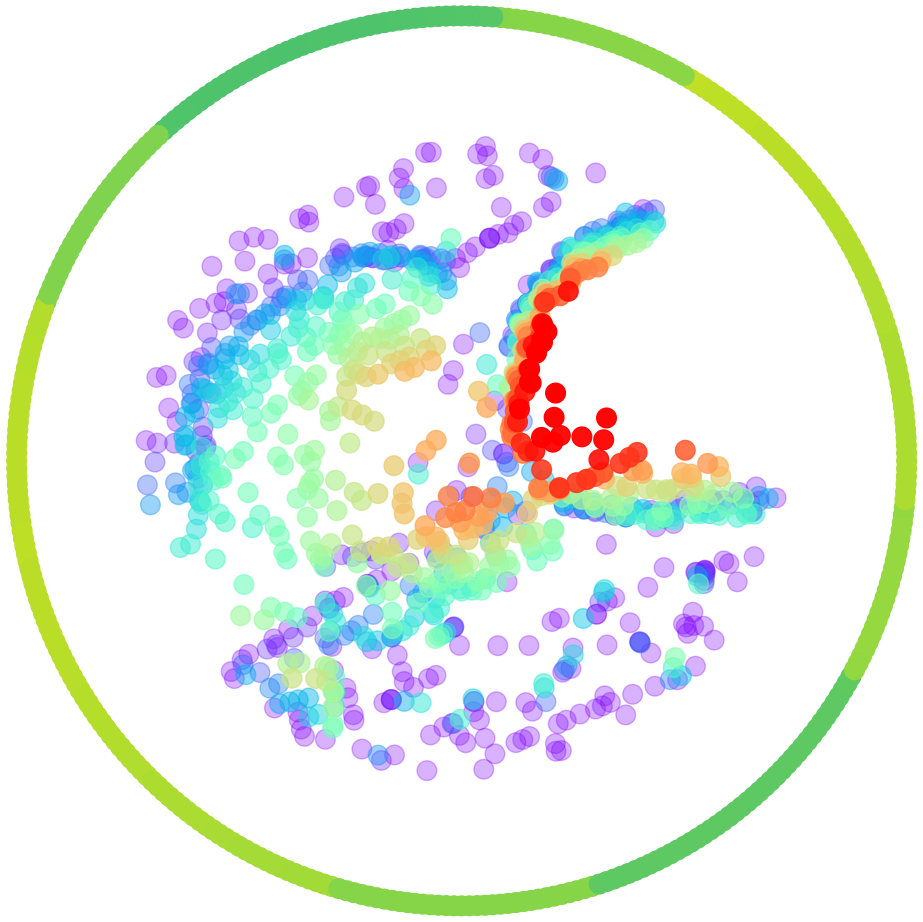}
        \captionsetup{skip=0pt}
        \caption{}
    \end{subfigure}
    \begin{subfigure}[t]{0.23\linewidth}
        \centering
        \includegraphics[width=\linewidth]{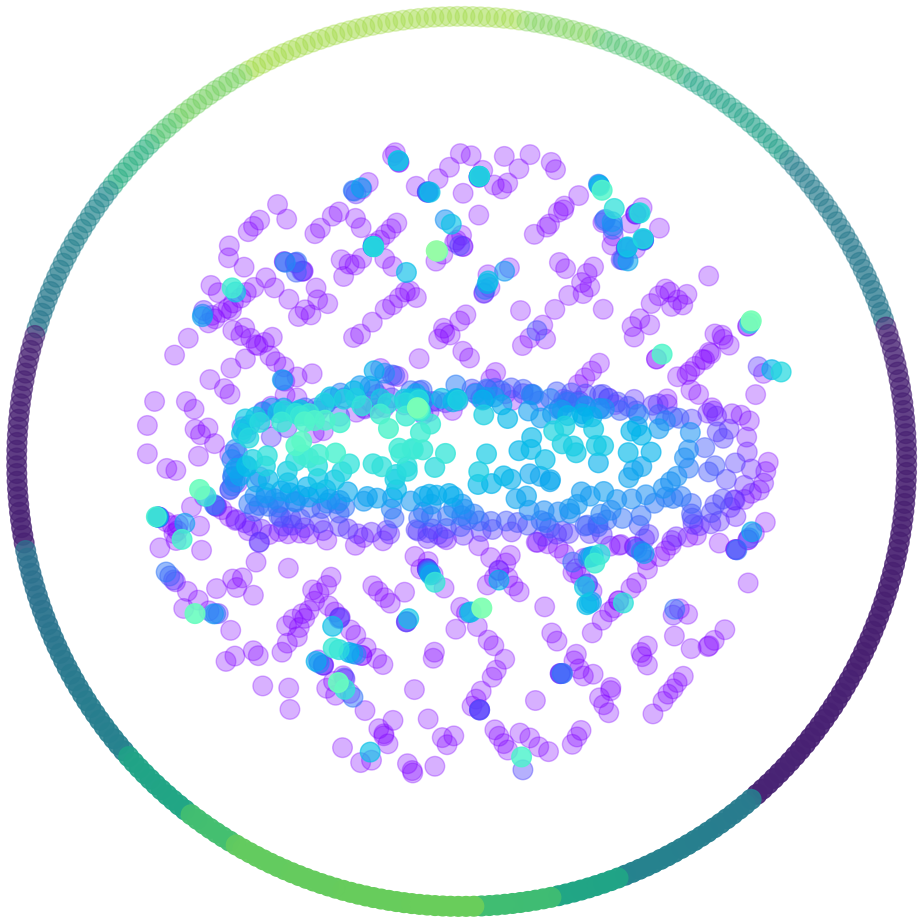}
        \captionsetup{skip=0pt}
        \caption{}
    \end{subfigure}
    \begin{subfigure}[t]{0.23\linewidth}
        \centering
        \includegraphics[width=\linewidth]{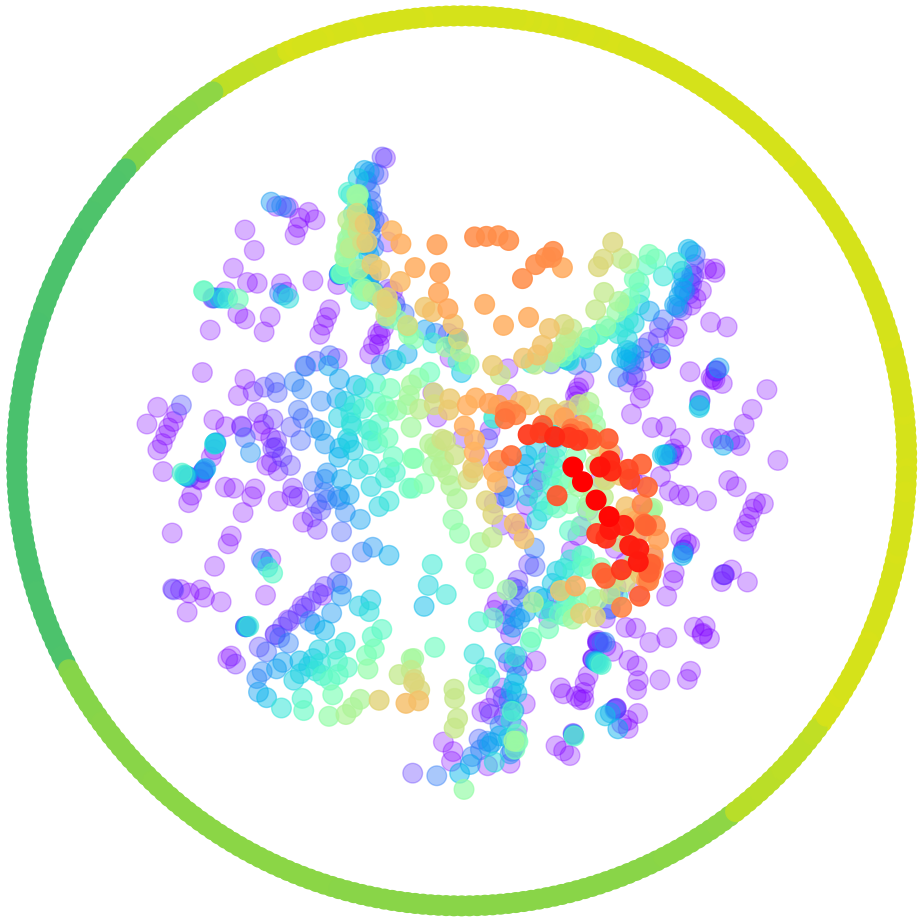}
        \captionsetup{skip=0pt}
        \caption{}
    \end{subfigure}
    \begin{subfigure}[t]{0.23\linewidth}
        \centering
        \includegraphics[width=\linewidth]{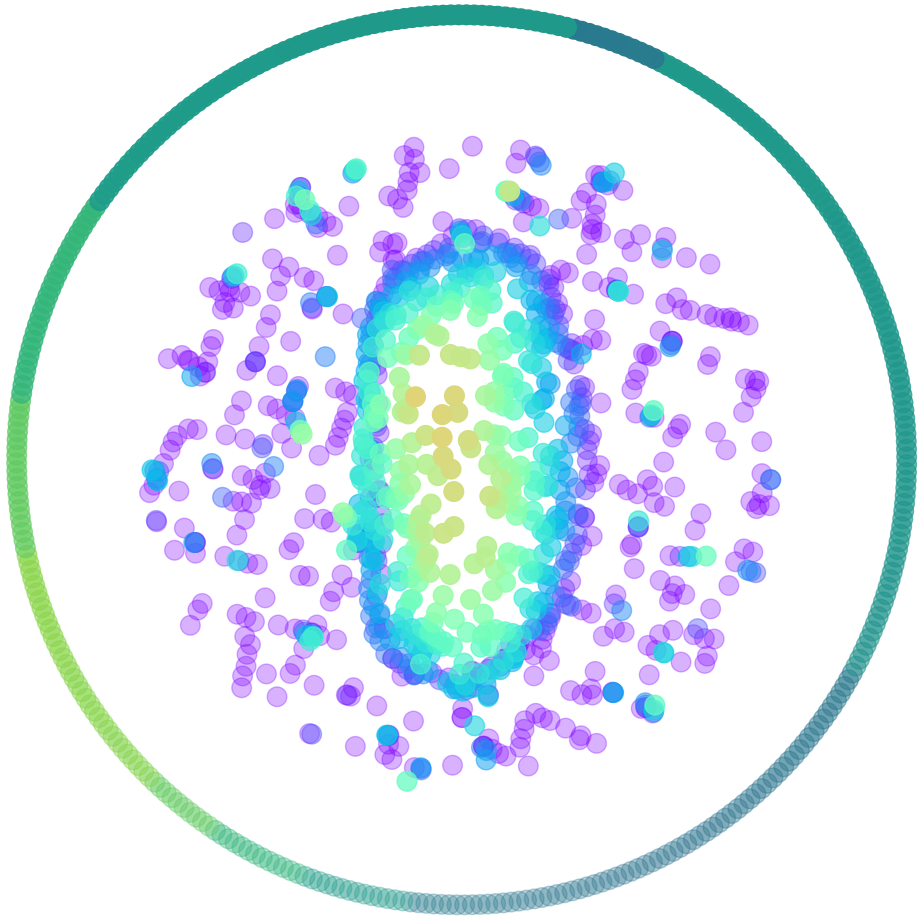}
        \captionsetup{skip=0pt}
        \caption{}
    \end{subfigure}
    \begin{subfigure}[t]{0.23\linewidth}
        \centering
        \includegraphics[width=\linewidth]{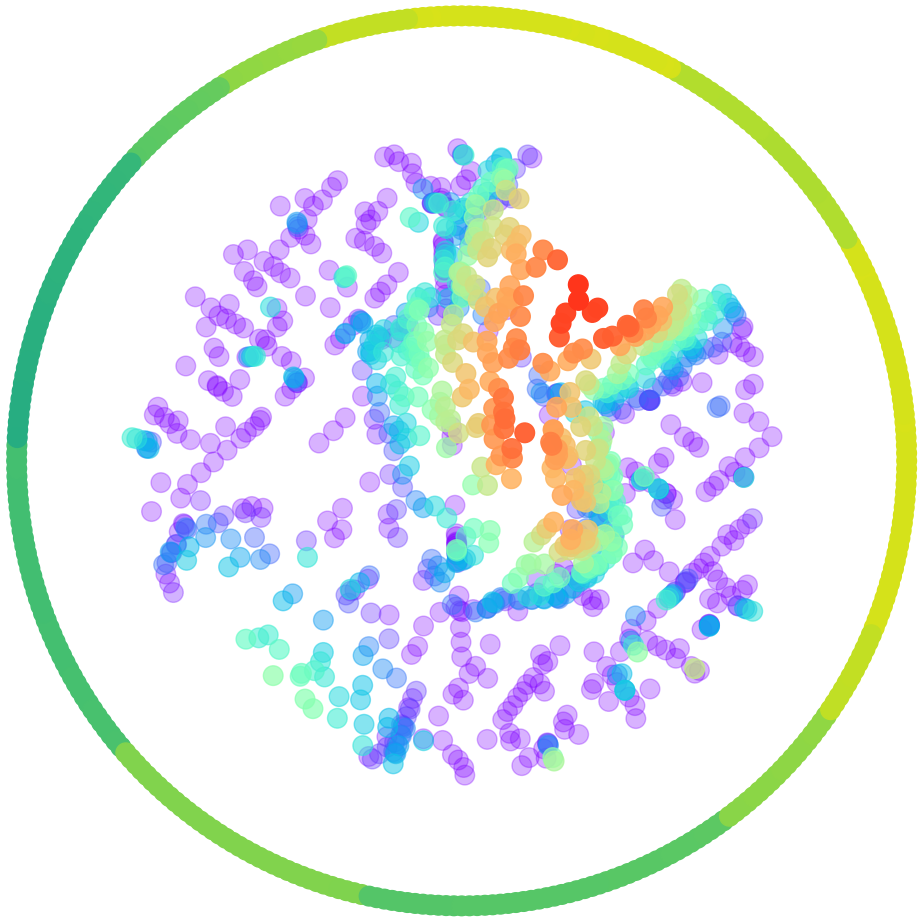}
        \captionsetup{skip=0pt}
        \caption{}
    \end{subfigure}
    \begin{subfigure}[t]{0.23\linewidth}
        \centering
        \includegraphics[width=\linewidth]{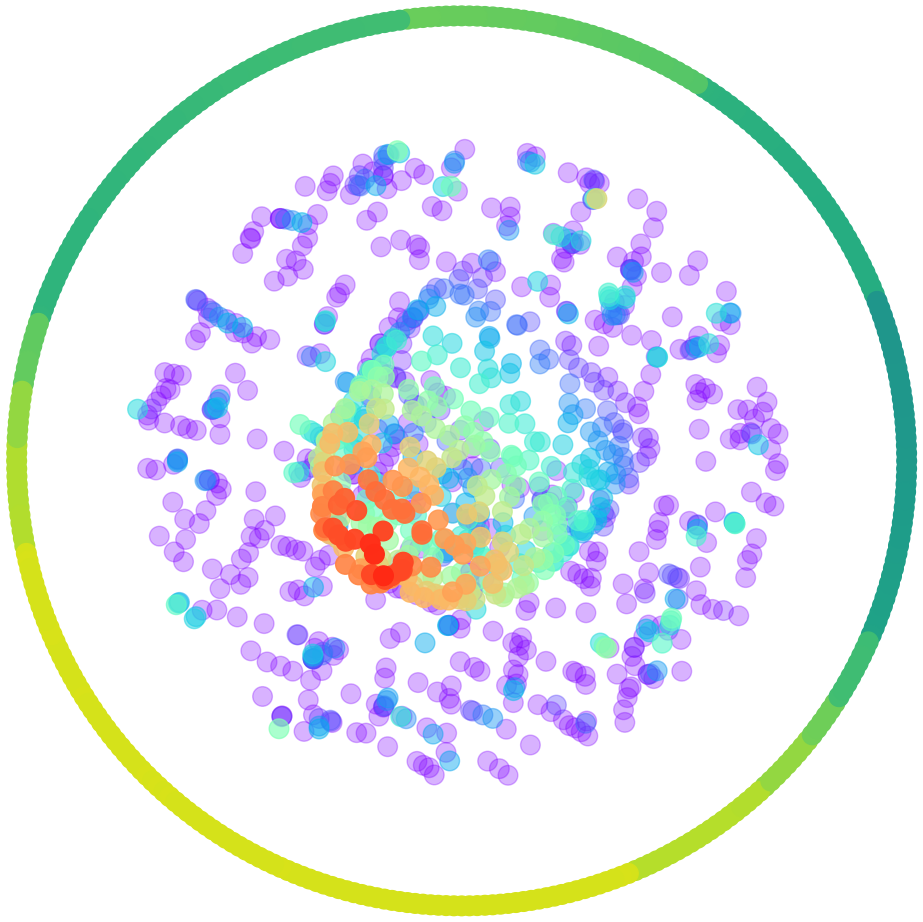}
        \captionsetup{skip=0pt}
        \caption{}
    \end{subfigure}
    \begin{subfigure}[t]{0.46\linewidth}
        \centering
        \includegraphics[width=\linewidth]{images/appendix/CVaR_colorbar.png}
        \captionsetup{skip=0pt}
    \end{subfigure}
    \begin{subfigure}[t]{0.46\linewidth}
        \centering
        \includegraphics[width=\linewidth]{images/appendix/elevation_colorbar.png}
        \captionsetup{skip=0pt}
    \end{subfigure}
    \vspace{-5pt}
    \caption{
        Example predictions of SPARTA. \textit{Rainbow} denotes point elevation, and \textit{Viridis} circles around the
        point clouds denote the CVaR at the corresponding angle of approach. The CVaR matches our intuition because it 
        indicates the obstacles are more risky from orientations with sharp edges and less risky from orientations with 
        smooth elevation change.
    }
    \label{fig:appendix:visualization}
    \vspace{-1.8em}
\end{figure}

In \cref{fig:appendix:visualization}, we visualize some example obstacles in our test set and
the CVaR predicted by our model with angle of approach $\phi \in \{0, 1, \dots, 359\}$.
The \textit{Rainbow} color map denotes the elevation of the point (Red denotes high elevation),
and \textit{Viridis} denotes the predicted CVaR (Yellow corresponds to high CVaR).
A trend we notice from the examples is that our model predicts lower CVaR for angles where the
obstacle has a smoother elevation change.
This is aligned with our intuition, further supporting the validity of our approach for identifying
risky obstacles and angles of approach.

% \end{document}

  % .bib

\end{document}